\documentclass[10pt,journal]{IEEEtran}

\ifCLASSOPTIONcompsoc
  \usepackage[nocompress,noadjust]{cite}
\else
  \usepackage{cite}
\fi

\ifCLASSINFOpdf
  \usepackage[pdftex]{graphicx}
  \graphicspath{{./pdf/}{./jpeg/}{./graphics/}{./ME/Figs/}{./photos/}}
\else
\fi

\usepackage{amsmath}
\usepackage{lipsum}
\usepackage{mathtools}
\usepackage{cuted}
\ifCLASSOPTIONcompsoc
  \usepackage[caption=false,font=footnotesize,labelfont=sf,textfont=sf]{subfig}
\else
  \usepackage[caption=false,font=footnotesize]{subfig}
\fi
\usepackage{ragged2e}

\usepackage{tikz}
\usetikzlibrary{shapes,arrows,scopes,positioning,calc,math,patterns,decorations.pathmorphing,decorations.markings,decorations.pathreplacing,spy,angles,quotes}
\usepackage{ctable}
\makeatletter
\def\thickhline{%
  \noalign{\ifnum0=`}\fi\hrule \@height \thickarrayrulewidth \futurelet
   \reserved@a\@xthickhline}
\def\@xthickhline{\ifx\reserved@a\thickhline
               \vskip\doublerulesep
               \vskip-\thickarrayrulewidth
             \fi
      \ifnum0=`{\fi}}

\usepackage{array}

\pretocmd\@bibitem{\color{black}\csname keycolor#1\endcsname}{}{\fail}
\newcommand\citecolor[1]{\@namedef{keycolor#1}{\color{red}}}
\makeatother

\newlength{\thickarrayrulewidth}
\usepackage{multirow}
\usepackage{makecell}
\usepackage{mcode}

\usepackage[inline]{asymptote}

\usepackage{siunitx}
\usepackage{nicefrac}

\let\oldfrac\frac
\makeatletter
\newcommand{\groupit}[1]{(#1)}
\newcommand{\nogroupit}[1]{#1}
\renewcommand{\frac}[2]{%
  \setbox\z@\hbox{$#1$}
  \setbox\tw@\hbox{$#2$}
  \ifdim\wd\z@>1em \let\groupornot@i\groupit\else\let\groupornot@i\nogroupit\fi
  \ifdim\wd\tw@>1em \let\groupornot@ii\groupit\else\let\groupornot@ii\nogroupit\fi
  \mathchoice
    {\oldfrac{#1}{#2}}
    {\groupornot@i{#1}/\groupornot@ii{#2}}
    {\groupornot@i{#1}/\groupornot@ii{#2}}
    {\groupornot@i{#1}/\groupornot@ii{#2}}
}
\usepackage{xcolor}
\usepackage[normalem]{ulem}

\usepackage{soul}
\usepackage{lscape}
\usepackage[utf8]{inputenc}
\usepackage{float}
\usepackage{rotating}
\usepackage{pifont}
\newcommand{\cmark}{\ding{51}}%
\newcommand{\xmark}{\ding{55}}%
\usepackage{enumitem}
\usepackage{hyperref}   
\usepackage{tabularray}

\begin{document}
\title{Optical Flow Based Detection and Tracking of Moving Objects for Autonomous Vehicles}

\author{MReza~Alipour~Sormoli$^{1}$,
Mehrdad~Dianati$^{1}$,~\IEEEmembership{Senior~Member,~IEEE,}
Sajjad~Mozaffari$^{1}$, and Roger Woodman$^{1}$%
\thanks{$^{1}$ Mehrdad Dianati holds part-time professorial posts at the School of Electronics, Electrical Engineering and Computer Science (EEECS), Queen’s University of Belfast and WMG at the University of Warwick. Other authors are with WMG, University of Warwick, e-mail: \{mreza.alipour, m.dianati, sajjad.mozaffari, r.woodman\}@warwick.ac.uk,  m.dianati@qub.ac.uk}%
\thanks{$^{\#}$Corresponding author: \tt\small mreza.alipour@warwick.ac.uk}
}

\markboth{IEEE TRANSACTIONS ON INTELLIGENT TRANSPORTATION SYSTEMS}{}%

\maketitle

\begin{abstract}
Accurate velocity estimation of surrounding moving objects and their trajectories are critical elements of perception systems in Automated/Autonomous Vehicles (AVs) with a direct impact on their safety. These are non-trivial problems due to the diverse types and sizes of such objects and their dynamic and random behaviour. Recent point cloud based solutions often use Iterative Closest Point (ICP) techniques, which are known to have certain limitations. For example, their computational costs are high due to their iterative nature, and their estimation error often deteriorates as the relative velocities of the target objects increase ($>$2 m/sec). Motivated by such shortcomings, this paper first proposes a novel Detection and Tracking of Moving Objects (DATMO) for AVs based on an optical flow technique, which is proven to be computationally efficient and highly accurate for such problems. \textcolor{black}{This is achieved by representing the driving scenario as a vector field and applying vector calculus theories to ensure spatiotemporal continuity.} We also report the results of a comprehensive performance evaluation of the proposed DATMO technique, carried out in this study using synthetic and real-world data. The results of this study demonstrate the superiority of the proposed technique, compared to the DATMO techniques in the literature, in terms of estimation accuracy and processing time in a wide range of relative velocities of moving objects. Finally, we evaluate and discuss the sensitivity of the estimation error of the proposed DATMO technique to various system and environmental parameters, as well as the relative velocities of the moving objects.

\end{abstract}


\IEEEpeerreviewmaketitle

\section{Introduction}
\IEEEPARstart{A}{ccurate}, reliable and fast perception of the surrounding environment is one of the most important technical challenges in the safe deployment of Autonomous/Automated Vehicle (AV) technologies. This problem includes the detection of surrounding objects and estimation of their states, i.e., their position and velocity. A particularly important element of this problem is associated with the Detection and Tracking of Moving Objects (DATMO), a.k.a. Moving Object Detection and Tracking (MODT) in some studies of the literature~\cite{sualeh2019dynamic}. 

 There is a wide range of DATMO techniques in the literature tailored for AVs that use camera perception sensors \cite{abbass2021survey,premachandra2020detection,sivaraman2013looking}, LiDAR perception sensors \cite{kusenbach2016new,borcs2017instant,steyer2018grid} and Radar perception sensors \cite{hutchison2010traffic}. LiDAR perception sensors are particularly popular for AVs as they inherently provide a wide field of view (FOV) and robust point clouds that can be used for highly accurate range estimations. Therefore, in this study, we focus on developing a DATMO technique primarily tailored for LiDAR perception sensors. However, we believe such a technique can be easily adapted to any type of perception sensor, such as depth cameras, that provides a point cloud output.

\begin{table*}[t]
\centering
\caption{A brief review of the LiDAR-based DATMO methods in the literature. \\ l: Learning-based, icp: interactive closest point method, g: Grid-based representation}
\label{table: review}
\begin{tabular}{lm{1.75cm}m{2.4cm}m{4cm}m{4cm}}
\toprule
\multicolumn{2}{c}{\textbf{Category}} & \multicolumn{1}{c}{\textbf{References}} & \multicolumn{1}{c}{\textbf{Advantages}} & \multicolumn{1}{c}{\textbf{Disadvantages}} \\ 
\midrule
\multirow{2}{*}{\parbox{2cm}{Detection-Based Tracking}} & Model-Based & \cite{sualeh2019dynamic}$_l$ \cite{ye2016object} & 
\begin{itemize}
\item independent detection and tracking
\item fusing different senors in detection level 
\end{itemize} 
& 
\begin{itemize}
\item tracking performance relies on detection/classification
\item difficult to detect and track objects with unknown geometries
\end{itemize} \\ \cline{2-5} 
 & Model-Free & \cite{spinello2010multiclass} \cite{douillard2008laser}  \cite{himmelsbach2008lidar}$_l^g$ \cite{shi2020pv}$_l$ \textcolor{black}{\cite{wang2022detection}$^g$}& 
 \begin{itemize}
\item tracking object with different geometries
\end{itemize} 
 & 
 \begin{itemize}
\item more false-negatives (FNs) in detection and tracking 
\end{itemize} 
 \\ 
\hline 
\multirow{2}{*}{Direct Tracking} & Model-Based & \rule{0pt}{10pt} \cite{petrovskaya2009model} \cite{an2020novel} \cite{steyer2019grid}$^g$  \cite{zhang2017efficient}& 
\begin{itemize}
\item sensor's physical model is considered $\to$ more accurate DATMO
\item geometry of the objects is tracked $\to$ accurate state estimation
\end{itemize}
& 
\begin{itemize}
\item tracking performance decreases for objects with different shapes  
\end{itemize} 
\\ \cline{2-5}
 & \begin{tabular}[c]{@{}l@{}} Model-Free\\ (Point-Based)\end{tabular} & \cite{steyer2018grid}$^g$ \cite{asvadi2015detection}$^g$ \cite{kaestner2012generative} \cite{lee2020moving}$_{icp}$ \cite{lee2022moving}$_{icp}$ \cite{moosmann2013joint}$_{icp}$ \cite{dewan2016motion}$_{icp}$ \cite{gross2019alignnet}$_{l}$ \cite{kim2021online}  \textcolor{black}{\cite{li2023high}$^g_{icp+l}$}& 
 \begin{itemize}
\item tracking all scanned points
\item DATMO performance doesn't rely on geometric shape
\end{itemize}
 & 
 \begin{itemize}
\item high computational cost
\item higher the relative velocity for moving objects $\to$ lower tracking performance
\end{itemize}
 \\ 
 \bottomrule
\end{tabular}
\end{table*}

When it comes to designing a DATMO technique, minimising its processing cost/time and estimation error are two significant challenges. If an AV uses a LiDAR perception sensor, the velocity of the surrounding moving objects can be calculated by corresponding two consecutive point cloud scans. In traditional approaches, an object detection function is used as the first processing stage to identify objects in two consecutive LiDAR scans. Then, a tracking algorithm is applied to compute the velocity for the objects of interest \cite{ye2016object}. Although these techniques are computationally efficient, their accuracy depends on the accuracy of the underlying object detection algorithms. While enhancing various elements of the techniques, for example, by utilising the geometric models of the moving objects in the detection process, can improve the performance of this category of DATMO techniques, such techniques do not perform well in many scenarios \cite{lee2020moving}. For example, if a vehicle travels at the speed of 90 km/hr on a highway, the lateral velocity estimation error of such techniques can exceed 2 m/sec. This is not regarded as an acceptable input to the planning modules that determine cut-in/cut-out intentions of vulnerable road users \cite{lee2022moving}. 

The problem can be exacerbated because the geometric models of the moving objects can significantly vary for various road users. This can have a direct impact on the estimation errors of the aforementioned DATMO techniques. To address these problems, a different category of DATMO techniques has emerged in the literature \cite{moosmann2013joint,dewan2016motion,gross2019alignnet,kim2021online}. These techniques often use point cloud registration algorithms such as Iterative Closest Points (ICP) \cite{lee2022moving} and track all moving points in the point cloud \cite{moosmann2013joint}; hence, they are more accurate in velocity estimation. However, these DATMO techniques are computationally expensive because of their iterative nature \cite{arnold2021fast}. Furthermore, the performance of the underlying point cloud registration methods can deteriorate when the deviation between two consecutive point cloud scans increases. For instance, if the relative speed of the Ego Vehicle (EV) and a target object of interest is high (e.g., 12 km/hr), the dislocation of the consecutive LiDAR scans is usually large, which can result in a large error in ICP-based point cloud registration algorithms ~\cite{lee2020moving}. This can result in poor performance of the latter category of the DATMO techniques. 


Motivated by the above challenges, in this paper, a novel DATMO technique is proposed for AVs that use LiDAR perception sensors. The proposed technique is inspired by \textit{optical flow} algorithm~\cite{farneback2003two}. In our approach, the 3D LiDAR scans are initially converted to 2.5D motion grids (Fig.~\ref{fig:conversion2grid}) inspired by\cite{asvadi2015detection}; then, the 2D velocity of each cell in the grid is estimated by comparing two consecutive LiDAR scans. In the next step, a series of grid mask filters such as temporal and rigid-body continuity filters are applied to eliminate false positive detection. The LiDAR points are classified based on their velocity vectors, and each class is associated with a moving object. Finally, a Kalman Filter is used to track the velocity and position of the detected moving objects, considering their dynamic model. \textcolor{black}{The main contributions of this paper are summarized as follows:
\begin{itemize}
    \item Adopting optical flow technique to process the 3D point cloud data instead of complex ICP algorithms. This is used to generate a grid-based velocity vector field representing a dynamic driving environment. 
    \item Introducing a two-layer filter applied to the velocity vector field eliminating the false positives and erroneous vectors. These filters are designed based on the spatial continuity of the vector field (rigid-body assumption) and temporal propagation to improve the estimation performance results.
    \item Introducing novel error model for velocity estimation as a function of a configuration set for target vehicles (TVs) w.r.t the ego vehicle (EV). This offers further insights to be incorporated into the downstream modules such as motion planning/prediction in the autonomous vehicle framework.
\end{itemize}
}
The performance of the proposed technique compared to the ICP-based methods, such as the one in~\cite{lee2020moving}, \textcolor{black}{\cite{li2023high}, model-free \cite{wang2022detection} and model-based indirect tracking methods \cite{zhang2017efficient}}, is evaluated in two steps. First, the compared DATMO techniques are evaluated on a synthetic dataset generated in MATLAB scenario designer, where various driving contexts are considered. In the next step, the KITTI tracking dataset from real driving scenarios \cite{Geiger2012CVPR} is used. Each one of the synthetic and KITTI datasets serves a different purpose in our study. The synthetic dataset enables generating a wide range of driving scenarios and various target vehicles (shape, velocity, dimension, etc.), which is not practically feasible in real-world data collection campaigns. The flexibility of this type of dataset becomes even more important when it comes to analysing error sensitivity to different factors that are easy to change or sweep in synthetically generated scenarios. On the other hand, testing the proposed DATMO techniques with data collected by contemporary sensors in real-world conditions helped us validate its performance in the real world. Comparing the estimation error distribution shows that the proposed DATMO outperform the state-of-the-art in both speed and yaw angle estimation. Moreover, the computational cost (without parallel calculations) shows improvements of about $10\%$, whereas parallelising the proposed method is easily available and could improve this metric even more significantly. The proposed error sensitivity analysis also revealed a meaningful correlation between the configuration of the TV and estimation error from which the researcher could benefit to develop motion planning/prediction algorithms.

\begin{figure*}[t]
\centering
\includegraphics[width= 0.85\linewidth]{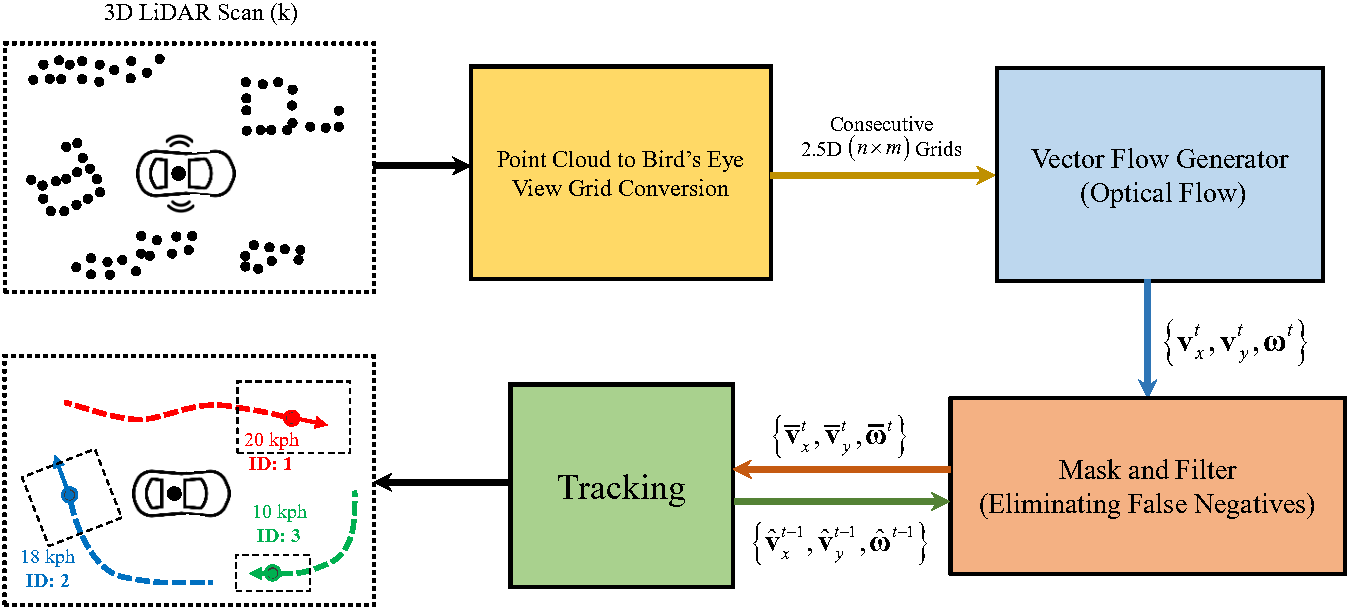}
\caption{High-level schematic system diagram of optical flow based DATMO for AVs. ${\bf{\omega }}$, ${\bf{v_x }}$, and ${\bf{v_y }}$ all are ${n \times m}$ matrices. The same colour code (blocks and signals) is used to expand and explain in different sections}
\label{fig:diagram}
\centering
\end{figure*}

The rest of the paper is organized as follows. An overview of the existing related work in the literature is described in Section \ref{sec: related works}. The system model and problem formulation are given in Section \ref{sec: problem definition}. The proposed DATMO method is explained in Section \ref{sec: proposed}. The performance evaluation methodology and results are discussed in Sections \ref{sec: experiment} to \ref{sec: discussion}. Finally, the key findings and conclusions of the study are given in Section.\ref{sec: conclusion}.

\section{related works}
\label{sec: related works}
In order to review available point cloud based DATMO/MODT approaches in the literature, in this paper, they are categorized into two main classes: 1) \textit{detection-based methods} also known as \textit{traditional methods} \cite{petrovskaya2012awareness}; 2) \textit{direct tracking methods} which is further divided into \textit{model-based} and \textit{point-based} approaches. 


\subsection{Detection-Based Tracking}
The \textit{detection-based} or indirect algorithms track the abstracted objects, patterns, bounding boxes, or clusters \cite{spinello2010multiclass,douillard2008laser} by applying different filters such as variants of the Kalman filter or particle filter. Therefore, the tracking performance for these methods relies on both classification algorithms (or pattern recognition) and filter structure \cite{ye2016object}. There is a great number of research aimed to improve the object tracking task by developing enhanced classification/clustering steps (before tracking) using learning-based \cite{himmelsbach2008lidar, shi2020pv} or geometric model estimation \cite{sualeh2019dynamic} algorithms, but all are still classified under the first category of DATMO methods. \textcolor{black}{There are other studies in this category, such as \cite{wang2022detection}, focused on reducing the computational complexity by applying the classification and subsequent tracking only on the moving point}. In a \textit{detection-based} approach the detection and tracking steps are independent and various sensor data could be used in the detection algorithm without changing the tracking part. 

\subsection{Direct Tracking}
In the direct tracking methods, the sensor model and/or object's geometric model is used to estimate corresponding points in space without prior detection. This method could be further divided into model-based and model-free (point-based) approaches.  

In \textit{model-bases direct tracking} DATMO algorithms, prior knowledge about the geometric shape and dynamic model of the moving vehicles are used to track the states and the geometric shape of the objects \cite{petrovskaya2009model, an2020novel} without detecting the objects first \cite{ye2016object,zhang2017efficient}. Tracking the geometry helps to predict the dynamic properties with higher precision and discard tracked objects with strange shapes or geometry changes. However, the tracking accuracy declines for moving objects with different shapes and geometry like cyclists or pedestrians. 

The second subclass of the \textit{direct tracking} approach (\textit{point-based tracking}) is a geometric model free in which every point is tracked in consecutive LiDAR scans. However, these scanned point clouds could be used directly or represented in the form of 2D/3D grid space before being used in a \textit{grid-based} tracking algorithms \cite{asvadi2015detection}. The key advantage of the \textit{point-based} DATMO stems from the fact that there is no assumption about the geometric shape of the object, and the objects are classified/detected based on tracking corresponding scanned points on them. But tracking all points makes the computation process expensive and limits the method in terms of the maximum number of moving objects in a scene \cite{kaestner2012generative}. To overcome this challenge, before tracking scanned points they are divided into static and moving categories by generating a static obstacle map (SOM) \cite{lee2020moving} \textcolor{black}{or filter objects of interested with the help of deep learning methods~\cite{li2023high}.}. 

Point cloud registration (PCR) algorithms are widely used in model-free DATMO methods. After clustering point clouds in consecutive scans, corresponding clusters are detected and PCR algorithms such as interactive closest point (ICP) \cite{besl1992method} are applied to each set (two clusters from the same object in different time steps) to calculate precise relative motion \cite{lee2020moving, moosmann2013joint, dewan2016motion, gross2019alignnet, kim2021online, lee2022moving}. Although, low standard deviation of error has been reported for tracking velocity (0.4 m/sec) and orientation (1.81 deg) for the moving objects \cite{lee2020moving}, these methods suffer from a number of considerable drawbacks. First of all, the computational time is not deterministic and depends on the number of moving objects. Secondly, The performance of the ICP algorithm highly depends on the initial conditions and the performance deteriorates when the relative velocity of the moving objects (with respect to EV) increases. Finally, because the PCR algorithms are based on the iterative optimization process, parallelizing these algorithms is not simple and straightforward. Various methods reviewed in this section are summarized along with advantages/disadvantages in Table.\ref{table: review}.

\begin{figure}[b]
\centering
\includegraphics[width=1\linewidth]{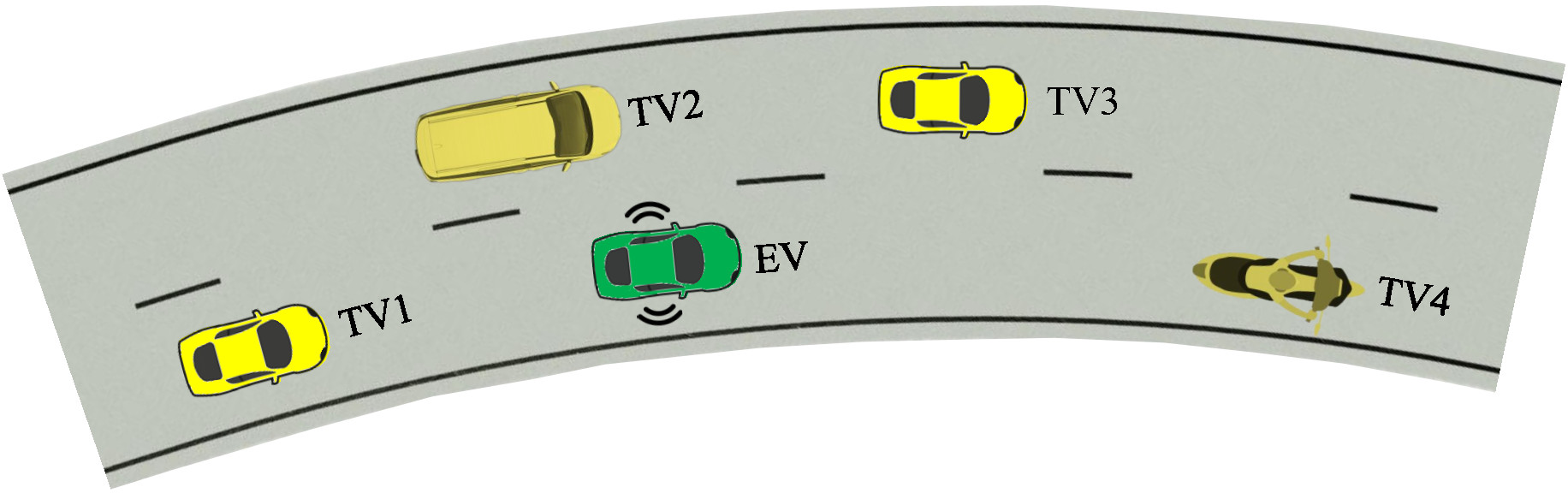}
\caption{Ego vehicle (EV) and target vehicles (TVs), including cars, vans, and bikers moving with different velocities}
\label{fig: road_segment}
\centering
\end{figure}

\section{System Model and Problem Definition}
\label{sec: problem definition}
We consider a system consisting of an EV equipped with LiDAR sensors and multiple target vehicles (TVs) such as cars, vans (or other large vehicles with different shapes), and bikers on a segment of a road as shown in Fig.~\ref{fig: road_segment}. The estimation of speed and direction of motion for all TVs is desired. 

\begin{figure}[t]
\centering
\includegraphics[width=1\linewidth]{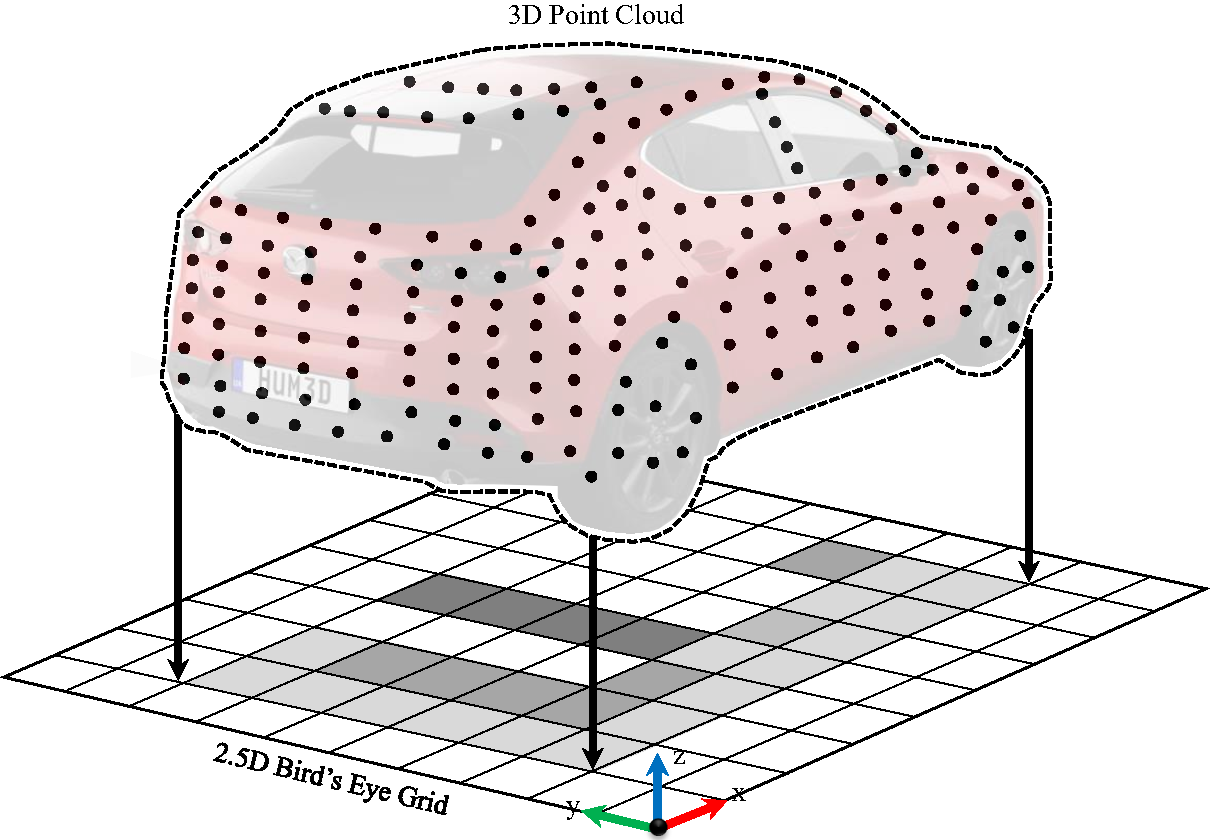}
\caption{3D point cloud conversion to 2.5 bird's eye view grid. Darker grids corresponds to higher value of ${G_{ij}}$, and ${G_{ij}} = 0$ in white cells.}
\label{fig:conversion2grid}
\centering
\end{figure}

As illustrated in Fig.~\ref{fig:diagram}, the input of the DATMO pipeline is a 3D point cloud (${P^l}$) generated by raycast LiDAR, and the desired output is a set of 2D velocity vectors ($\left\{ {{\bf{\hat v}}_1^t,{\bf{\hat v}}_2^t, \ldots ,{\bf{\hat v}}_o^t} \right\}$) each belongs to a unique track (trace of velocity vectors in a certain period of time). It should be noted that ``\textit{o}" is the number of moving objects at a time ``\textit{t}" (stationary objects are not included). Unlike \cite{lee2020moving}, in this study, we don't assume a small relative velocity for the moving targets. Moreover, the update rate of the LiDAR sensor is presumed 10 Hz, so, in order to avoid losing data, the proposed algorithm should be able to calculate the desired output (within a radius of 120 m) in less than 100 ms (regardless of the number of moving objects). However, we assumed that the ego vehicle and the surrounding objects move in the horizontal plane (xy in Fig.\ref{fig:conversion2grid}). Therefore the velocity in the vertical direction (z) is ignored and not reported in the estimated output. The estimated velocities are in the local coordinate system attached to the EV.

\section{Proposed Method}
\label{sec: proposed}
The proposed DATMO is illustrated using a block diagram of processes as shown in Fig.~\ref{fig:diagram}. The algorithm between input and output signals is divided into four main processes summarized as follows:   

\begin{enumerate}[label=(\Alph*)]
    \item 3D LiDAR sensor data are converted to 2.5D grayscale grid map.
    \item A velocity vector field generated using the optical flow algorithm.
    \item The false positive estimation are eliminated by a filtering mask calculated based on the continuum property of the velocity vector field for rigid-body motion
    \item Finally fusing all measured information and dynamic model in an Extended Kalman Filter (EKF)
\end{enumerate}

The core process is the optical flow calculation and the rest are either for preparing input data for this step or post-processing the generated vector flow for filtering and tracking the true positives. In the following subsections, each process in the pipeline is described further in detail.

\subsection{Point Cloud to Bird's Eye View Conversion Process}
The optical-flow algorithm is the main component of the proposed method, and this process requires 2D grayscale images to calculate the velocity vector field. However, the input data from LiDAR is 3D scattered point cloud. Similar to \cite{asvadi2015detection} a conversion block is utilized for mapping the point cloud input to a bird's eye view grid which is also known as 2.5D grid map (Fig.~\ref{fig:conversion2grid}). The input signal which is fed into the conversion process is a point cloud containing $L$ points each has three coordinates without intensity data, e.g. the $l^{th}$ point is represented by ${P^l} = \{ p_x^l,p_y^l,p_z^l\} $. The output of the conversion process (input of the vector flow generator) is a grayscale 2D image which is called 2.5D grid map in this paper. In other words, the output is a ${n \times m}$ matrix in which cells are normalized between 0 and 255. Each cell is referred to by an $ij$ pair where $i$ ($j$) is an integer between 1 and $N$ ($M$). Moreover, the centre of each cell in the grid has also a coordinate ($G_{ij}^x,G_{ij}^y$). Based on the grid's resolution, all cells have the same dimensions of $w$ and $h$ in $x$ and $y$ directions, respectively. A value is assigned to each cell of this grid space based on the height of the corresponding points in the point cloud data i.e. the points with the same x and y values (see Eq.~\ref{eq:gridCondition}). This value is calculated based on a linear combination of the mean and standard deviation of the height of the corresponding points projected on the horizontal plane. This concept has been illustrated in Fig.~\ref{fig:conversion2grid} and the value assigned to cell $ij$ is obtained as follows:

\begin{equation}
    {G_{ij}} = \frac{1}{{{h_{\max }}}}\left[ {a \cdot \mu \left( {P_z^g} \right) + b \cdot \sigma \left( {P_z^g} \right)} \right]
    \label{eq: gridValue}
\end{equation}

where, $a$ and $b$ are constant weight, and $h_{\max}$ is normalizing constant. $\mu(\cdot)$ and $\sigma(\cdot)$ are mean and standard deviation functions, respectively. Superscription $g$ is a set of integers ${1,2,..,L}$ that satisfies the following condition:

\begin{equation}
  \left\{ {\begin{array}{*{20}{c}}
{\left( {G_{ij}^x - {w \mathord{\left/
 {\vphantom {w 2}} \right.
 \kern-\nulldelimiterspace} 2}} \right)}& \le &{p_x^g}& < &{\left( {G_{ij}^x + {w \mathord{\left/
 {\vphantom {w 2}} \right.
 \kern-\nulldelimiterspace} 2}} \right)}\\
{\left( {G_{ij}^y - {h \mathord{\left/
 {\vphantom {h 2}} \right.
 \kern-\nulldelimiterspace} 2}} \right)}& \le &{p_y^g}& < &{\left( {G_{ij}^y + {h \mathord{\left/
 {\vphantom {h 2}} \right.
 \kern-\nulldelimiterspace} 2}} \right)}
\end{array}} \right.
\label{eq:gridCondition}
\end{equation}

\begin{figure}[b]
\centering
\includegraphics[width=1\linewidth]{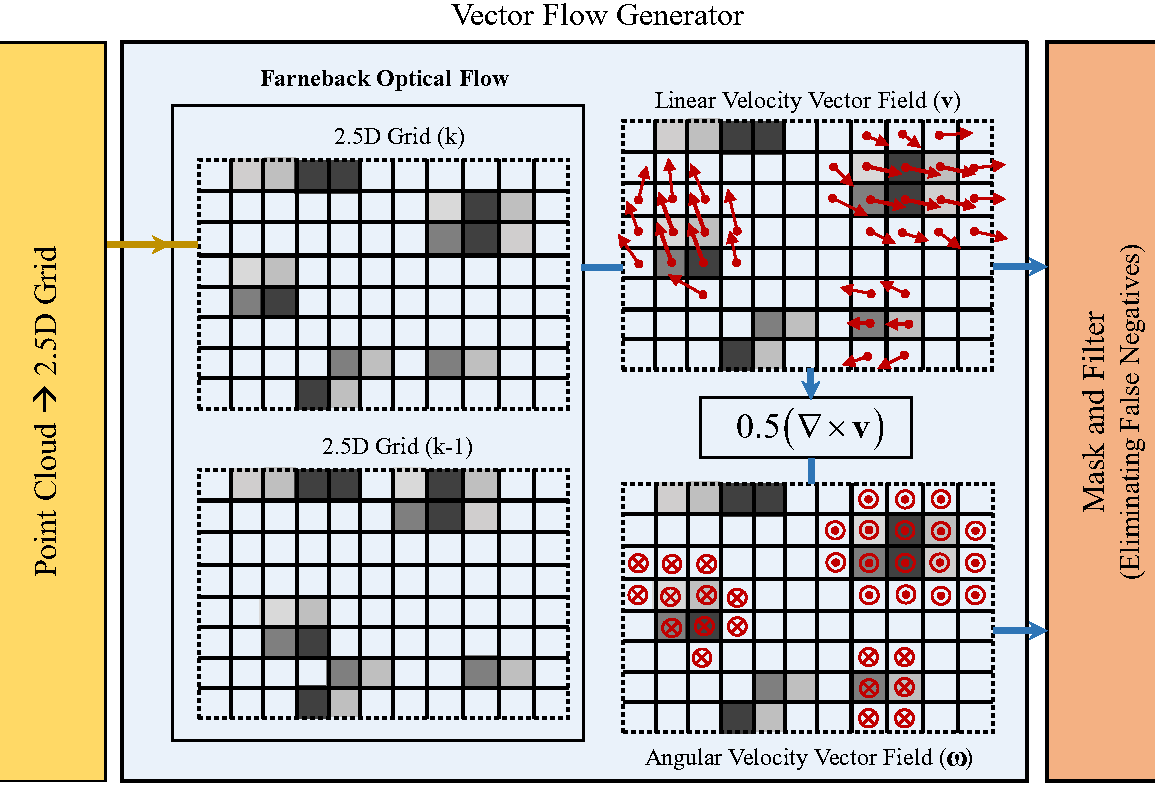}
\caption{Optical flow based velocity vector field generation process. Grayscale brightness refers to the occupied cells which contain LiDAR scanned points. $ \otimes $ and $ \odot $ show the angular velocity vectors perpendicular to the motion plane in $-z$ and $+z$, respectively. NOTE: for reading the system diagrams used in this paper the signals are expanded (rectangles with dashed line frame) to illustrate data that is carried between processing blocks.}
\label{fig:vectorFlow}
\centering
\end{figure}

Based on this definition, $G_{ij} = 0$ means that there is no point in the point cloud corresponding to the cell $ij$ above ground. Lower values (relative to a threshold) show that the points are from ground \cite{asvadi2015detection}. On the other hand, higher values are assigned for the points on vertical planes (bigger height standard deviation) or horizontal plane with high z components (bigger average height).  

\subsection{Motion Vector Flow Generation using Farneback Optical Flow Algorithm}
\label{sec: motion vector}

A dense optical flow algorithm is used to calculate the velocity (linear and angular) for each cell of the grid map generated in the previous step using two consecutive frames of the converted 2.5D grayscale grid map (Fig.\ref{fig:vectorFlow}). The output signal of this process carries three ${n \times m}$ matrices including linear and angular velocities of each cell in the grid. Two matrices for linear velocities in x and y directions (${\bf{v_x }}$ and ${\bf{v_y }}$, respectively), and one for angular yaw rate in the z-direction (${\bf{\omega }}$). 

There are several optical flow algorithms available in the literature for estimating the per-pixel motion between two consecutive images \cite{fortun2015optical}. In this study, we need dense vector flow (not sparse) to calculate the velocity of all occupied cells with high accuracy and low computational cost. Although any dense optical flow algorithm could be used in the proposed DATMO framework, the well-known Gunar-Farneback optical flow generator~\cite{farneback2003two} has been used here. This algorithm satisfies the requirements (accuracy-cost trade-off) more efficiently compared to other methods \cite{tanas2017comparison}. However, the Farneback algorithm employs expanded polynomial transformation of adjacent cell's brightness to estimate the dense velocity distribution for each grid cell~\cite{farneback2003two}, and this may cause estimating non-zero velocities for unoccupied cells in the neighbourhood of the occupied cells. This challenge is addressed in the next part (filtering and masking process).

\begin{figure}[!t]
\centering
\includegraphics[width=0.7\linewidth]{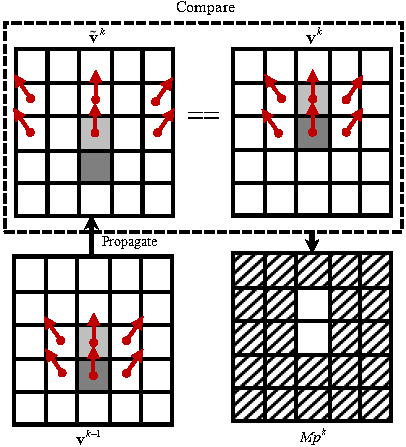}
\caption{Vector field propagation mask in time step $k$}
\label{fig:mask_p}
\centering
\end{figure}

The optical flow algorithm calculate linear velocity distribution, however, the angular velocity is also required for accurate state tracking of the vehicles (sec.\ref{sec: tracking}). Based on vector field theory~\cite{urwin2014advanced} and rigid body assumption for each moving object in the scene, the angular velocity of each cell (in $z$ direction) is obtained by the Eq.~\ref{eq: angVel}. (see~\cite{casey1983treatment} and Appendix).

\begin{figure*}[t]
\centering
\includegraphics[width= 1\linewidth]{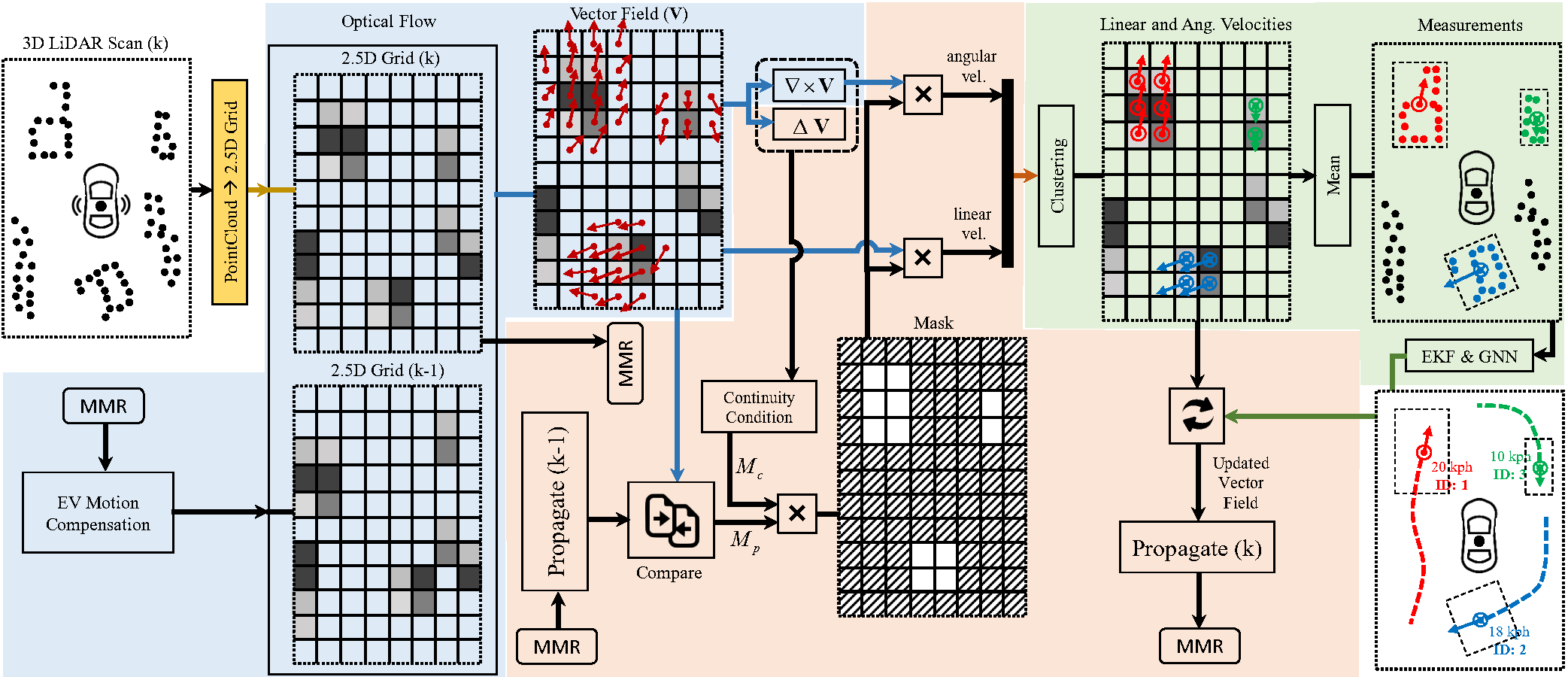}
\caption{Expanded schematic system diagram of optical flow based DATMO for AVs. NOTE: for reading the system diagrams used in this paper the signals are expanded (rectangles with dashed line frame) to illustrate data that is carried between processing blocks. And, \textbf{MMR} stands for memory.}
\label{fig:diagram2}
\centering
\end{figure*}

\begin{figure}[!b]
\centering
\includegraphics[width=1\linewidth]{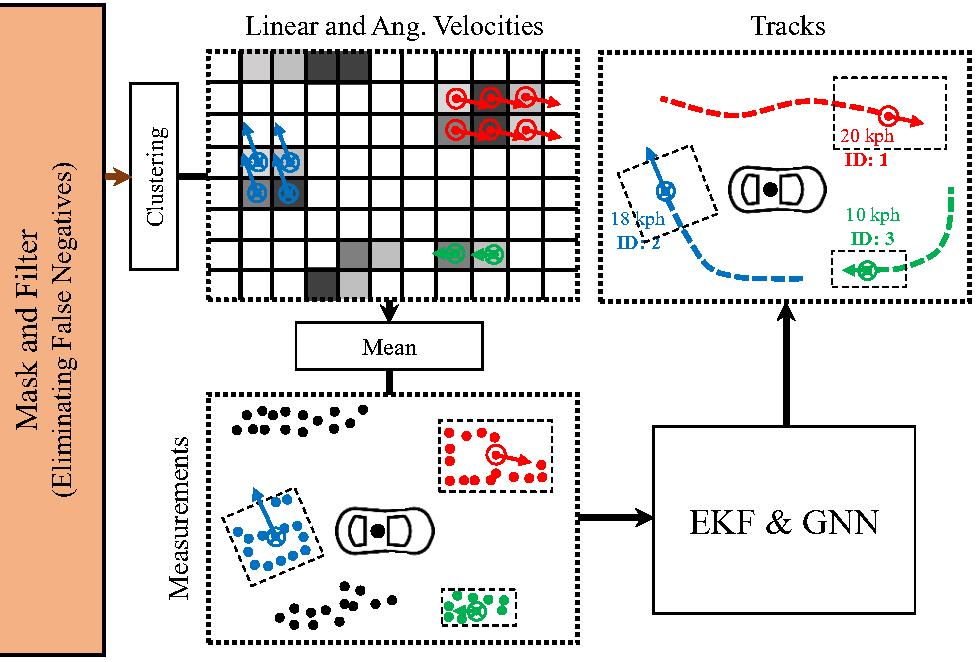}
\caption{Tracking process system diagram.}
\label{fig:tracks}
\centering
\end{figure}

\begin{equation}
   {\bf{\omega }} = 0.5\left( {\nabla  \times {\bf{v}}} \right)
\label{eq: angVel}
\end{equation}

\noindent Where $\bf{v}$ is the linear velocity vector field, and $\nabla$ is the curl operator. 

Therefore, the output of the vector flow generation process includes angular velocity in $z$ direction, in addition to 2D linear velocity in $x$ and $y$ directions. 

\subsection{Masking and Filtering the vector field}
\label{sec: mask}
Due to the dense nature of the vector field obtained by the Farneback optical flow algorithm, the generated vector field contains false positive values for cells which are not occupied or do not belong to moving objects (static). The masking process is to filter out undesirable false positives and provides the final estimated velocity vectors, so the output of this step is a subset of its input. In this section, the mask is obtained in two steps and prepare the final vector filed for the tracking process. 
 
\subsubsection{Vector Field Propagation Mask}
The second masking layer for the vector field is based on temporal filtering which is called \textit{propagation} in this study. Propagation of the vector filed in time step $k$ is obtained by changing the $(x,y)$ position of the velocity vectors in the 2D plane according to the linear velocity values in time step $(k-1)$. The propagation is calculated using Eq.~\ref{eq: propagation}.

\begin{equation}
\begin{array}{l}
\begin{array}{*{20}{l}}
{{\bf{\tilde v}}_{i'j'}^k}& = &{{\bf{\hat v}}_{ij}^{k - 1}}
\end{array};\\
\left\{ {\begin{array}{*{20}{l}}
{i'}& = &i& + &{\left\lfloor {{\hat v}_x^{k - 1}dt + \frac{w}{2}} \right\rfloor }\\
{j'}& = &j& + &{\left\lfloor {{\hat v}_y^{k - 1}dt + \frac{h}{2}} \right\rfloor }
\end{array}} \right.
\end{array}
\label{eq: propagation}
\end{equation}

where ${{\bf{\tilde v}}}$ is propagated vector field, and $dt$ is time increment. The value inside $\left\lfloor  \cdot  \right\rfloor$ is rounded down to the nearest integer. 

As shown in Fig.~\ref{fig:mask_p}, the propagated vector field of time step $(k-1)$ is compared with the vector field calculated in the same time step and the masking matrix ($Mc$) is obtained using Eq.~\ref{eq: masking_p}:

\begin{equation}
(Mp)_{ij}^k = \left\{ {\begin{array}{*{20}{l}}
1&{if}&{\left\| {{\bf{\tilde v}}_{ij}^k - {\bf{\hat v}}_{ij}^k} \right\| \le {\alpha _p}}\\
0&{}&{{\rm{otherwise}}}
\end{array}} \right.
\label{eq: masking_p}
\end{equation}

In this equation, $\alpha_p$ is a constant threshold close to zero. The final masking boolean matrix is calculated by multiplying two masks calculated in two layers: $Mask = Mc \times Mp$. Applying this filtering mask to the vector field at each time step filters out undesirable false positive vectors. 

\subsubsection{Rigid-Body Continuity Mask}
In this study, it has been assumed that the moving objects are rigid i.e different parts of a single object have zero relative motion. Therefore, linear ($\bf{v}$) and angular ($\bf{\omega}$) velocity vector fields should satisfy the continuity conditions of Eq.~\ref{eq: continuity}~\cite{urwin2014advanced}:

\begin{equation}
   \left\{ {\begin{array}{*{20}{l}}
{\Delta   {\bf{V}}}& = &0\\
{\nabla   \left( {\nabla  \times {\bf{V}}} \right)}& = &0
\end{array}} \right.
\label{eq: continuity}
\end{equation}

The first part of Eq.~\ref{eq: continuity} is for continuity in the linear velocity i.e the objects cannot tear apart nor implode, while the second part refers to the fact that all points on a single object should rotate with the same angular velocity. The results of both operations are 2D matrices, so, the estimated values for the cells that do not satisfy the condition (not exactly equal to zero but below a certain threshold $\alpha_{cont}$) are set to zero. The resulting mask from this procedure is referred to by $(Mc)$ in the rest of the text. 
\subsection{Tracking}
\label{sec: tracking}
The resulting vector field is used to detect moving objects and estimate their velocity. The tracking process output is the final estimated state of the objects (linear and angular velocities) augmented with a unique ID. As illustrated in Fig.\ref{fig:diagram}, $\bar x$, $\hat x$. and $\tilde x$ are masked, estimated, and propagated values of $x$ variable, respectively, while the superscription shows the time step for the variables. An Extended Kalman Filter (EKF) is designed to use vector field data as measurements and the dynamic model of Eq.\ref{eq:dynamicModel} (constant linear/angular acceleration) as the prediction model to estimate the state ($X_n$) of the moving objects. Every estimated position and velocity is assigned to either an existing or new track with a unique ID via Global Nearest Neighbour (GNN) \cite{lee2020moving}.

\begin{equation}
   \begin{array}{l}
{{\dot X}_n} = f\left( {{X_n},U} \right)\\
\\
\left[ {\begin{array}{*{20}{l}}
{{{\dot x}_n}}\\
{{{\dot y}_n}}\\
{{{\dot \theta }_n}}\\
{{{\dot v}_n}}\\
{{{\dot \omega }_n}}
\end{array}} \right] = \left[ {\begin{array}{*{20}{l}}
{{v_n}\cos {\theta _n} - v + {l_n}{\omega}}\\
{{v_n}\sin {\theta _n} - {l_n}{\omega}}\\
{{\omega _n} - \omega }\\
{{k_a}}\\
{{k_\alpha }}
\end{array}} \right];\quad U = \left[ {\begin{array}{*{20}{c}}
v\\
\omega 
\end{array}} \right]
\end{array}
\label{eq:dynamicModel}
\end{equation}

\subsubsection{Measurements in EKF and Updating Tracks}
In the Kalman filter structure three measurements are used for each moving object i.e two linear velocities and one angular velocity in $z$ direction. As illustrated in Fig.\ref{fig:tracks}, these measurements are calculated by clustering masked velocity vector fields $\left\{ {{{{\bf{\bar v}}}_{\bf{x}}},{{{\bf{\bar v}}}_{\bf{y}}},{\bf{\bar \omega }}} \right\}$ provided by optical flow and taking the mean value of each cluster. In the proposed approach, Euclidean distance is utilized for clustering vectors, and mean position and velocity are fed into the EKF algorithm to estimate the state vector for each moving object using motion dynamics of Eq.\ref{eq:dynamicModel}.

All clustered points should be either assigned to an existing track or initialised on a new track. Similar to \cite{lee2020moving}, in our approach, the clusters are assigned to the predicted tracks via GNN. Each cluster is assigned to at most one track based on a 4D feature vector $\left[ {{x_m},{y_m},{\lambda _1},{\lambda _2}} \right]$ containing the mean position and shape of the cluster (independent of the orientation) in the motion plane. Two components in the features vector showing the shape of a cluster are eigenvalues of the covariance matrix of the points in a cluster $({{\lambda _1},{\lambda _2}})$. So, a cluster is assigned to a track if the Euclidean distance between their feature vectors is less than a threshold $\gamma$. 

\begin{figure}[!t]
\centering
\includegraphics[width=0.7\linewidth]{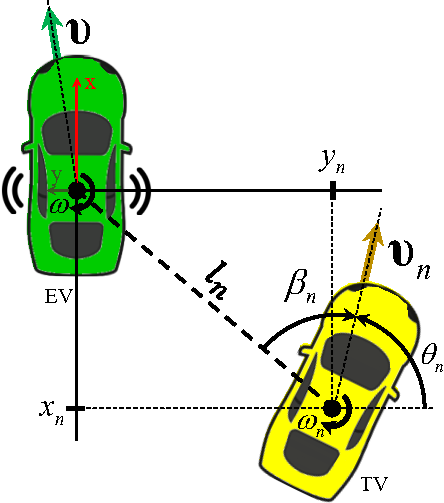}
\caption{Target vehicle's (TV) configuration (position, orientation, and velocity) relative to the ego vehicle (EV)}
\label{fig:dynamics}
\centering
\end{figure}

The final step in managing the tracks is confirming and/or deleting tracks. Every one of these two procedures is done by a 2D integer vector. A track is confirmed when $M_1$ number of measurements/detection is assigned to it in the last $N_1$ updates ($M_1<N_1$). And similarly, a confirmed track is deleted if in the last $N_2$ ($M_2<N_2$) consecutive updates, no measurement is assigned to it $M_2$ times. It should be noted that the coordinate system used in this section is attached to EV with a configuration shown in Fig.\ref{fig:dynamics}.

The interaction between different processes is depicted in the assembled system diagram of Fig.\ref{fig:diagram2}. This system diagram is a detailed version of Fig.\ref{fig:diagram}. 

\section{Performance Evaluation} 
\label{sec: experiment}
An experimental test is designed to evaluate and verify the performance of the designed DATMO algorithm. \textcolor{black}{Two main objectives are targeted in this section. Comparing the proposed DATMO with state-of-the-art (SOTA) methods, and obtaining an error model for estimation accuracy.} 

\textcolor{black}{First and foremost we statistically compare the performance of the DATMO method with SOTA geometric model-free approach (GMFA) developed in \cite{lee2020moving} which has been proven to be more efficient than the geometric model-based tracking (MBT) method proposed in \cite{cho2014multi}. However, the proposed method is further compared against SOTA model free \cite{wang2022detection} and model-based\cite{zhang2017efficient} direct tracking methods as well. The GMFA algorithm is regenerated and evaluated, while the quantitative performance of the other methods are obtained from \cite{wang2022detection} and \cite{zhang2017efficient}. Therefore, the later experiments are consistent with those specified in these studies.}   

\begin{figure*}[t]
\centering
\includegraphics[width= 0.98\linewidth]{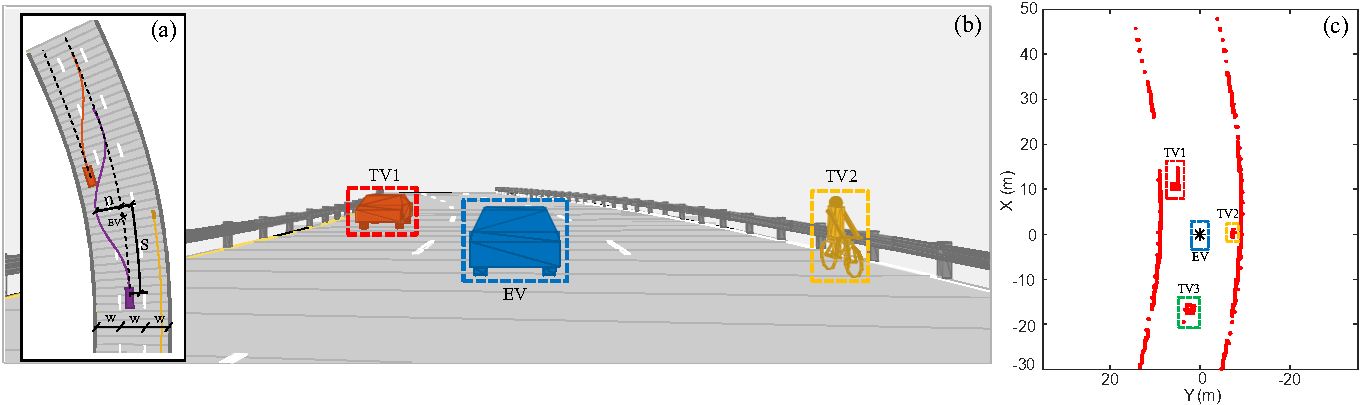}
\caption{Synthetic primary scenario generated in Matlab scenario designer: trajectories design parameters for different TVs (a), a 3D meshed object in scene (b), and bird's eye view scanned point cloud excluding ground points in EV coordinate system (c)}
\label{fig:synth_scenario}
\centering
\end{figure*}

\begin{figure*}[t]
\centering
\includegraphics[width= 1\linewidth]{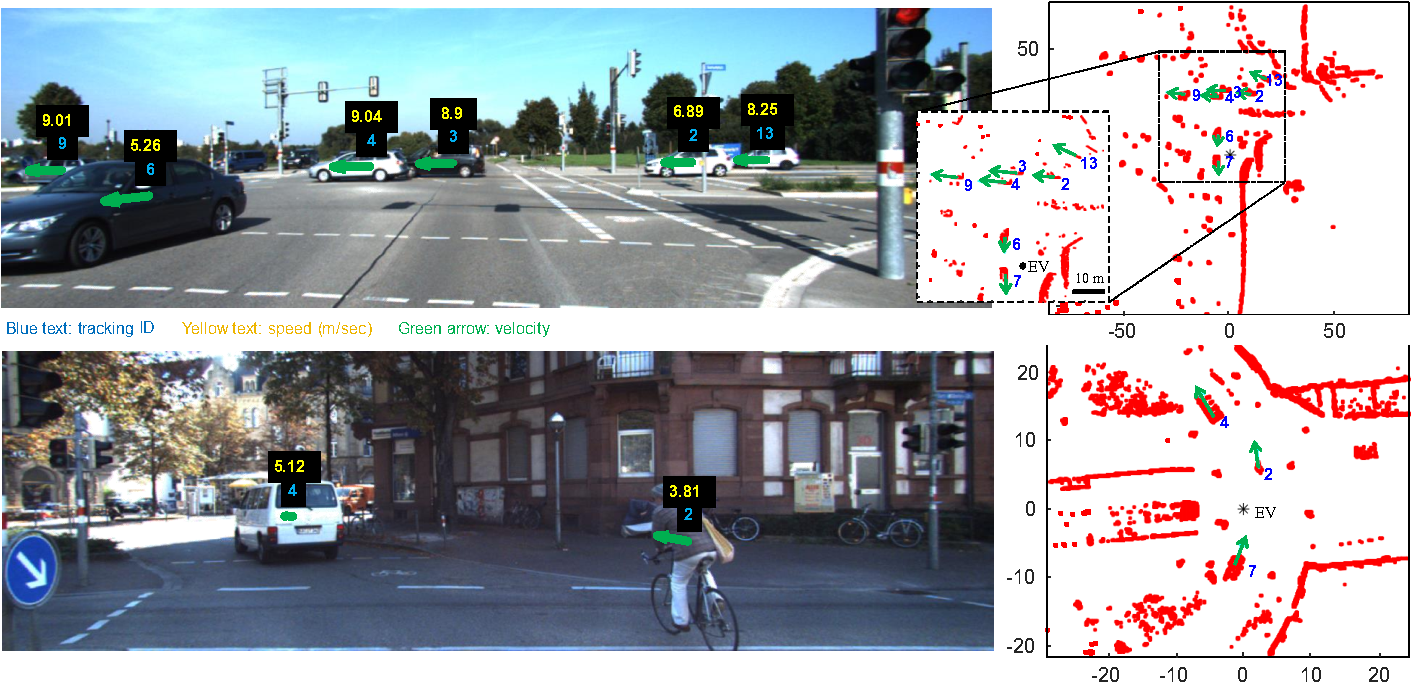}
\caption{Proposed DATMO algorithm result for KITTI dataset: tracking IDs and velocity vectors plotted on the front camera  image (left), and bird's eye view scanned point cloud excluding ground points (right)}
\label{fig:kitti_scenario}
\centering
\end{figure*}

In addition, we further investigate how the state estimation error changes as a function of the deriving environment i.e configuration of EV and TV. Regarding these objectives, the experimental evaluation is conducted in two different steps. Initially, synthetic data is generated to evaluate the algorithm in various custom situations, and in the next step, the algorithm is tested on a real-world dataset of KITTI.

\subsection{Datasets}

\subsubsection{Synthetic Data Generation and Simulation}
\label{sec: synthetic}
In order to evaluate the proposed method for estimating the state of the target vehicles in diverse possible configurations, generating a synthetic dataset is essential. In addition, the estimation error is calculated more accurately in the simulation compared with real-world datasets such as KITTI for which the ground truth of objects' velocity has not been provided directly. In this study, the TV's configuration space is defined by three variables (Fig.~\ref{fig:dynamics}): distance to EV ($l_n$), relative orientation ($\beta_n$), and relative velocity ($\Delta {{\bf{\upsilon }}_n} = {{\bf{\upsilon }}_n} - {\bf{\upsilon }}$). The aim is to design scenarios covering all possible configurations for investigating the meaningful relations between the estimation error and these three variables, in addition to assessing the estimation accuracy. Therefore, the flexibility in changing different configurations provided by synthetic datasets is another reason that justifies utilising this type of dataset. 

The driving scenario designer toolbox in MATLAB is used to generate synthetic scenarios and add a LiDAR sensor to collect point cloud data. As illustrated in Fig.~\ref{fig:synth_scenario}-(b), three different types of TVs are simulated in synthetic scenarios including sedan, van, and cyclist. Moreover, for considering EV motion effect completely, a nonzero curvature is considered to avoid zero yaw rate for EV. The LiDAR sensor parameters are adjusted according to what is used in the KITTI dataset collection sensor. The point cloud data from a simulated scene has been plotted in Fig.~\ref{fig:synth_scenario}-(c).

In order to cover all possible cases for the $n$-th target vehicle configurations ($\left\{ {{l_n},{\beta _n},\Delta {\upsilon _n}} \right\}$), each scenario contains a target vehicle (sedan, van, or cyclist) moving in the same multi-lane road in which EV moves in one of the lanes (with a speed of 20 m/s). TVs move with 10 different speeds (\textcolor{black}{10 to 40 with a step of 2 m/s}) in a lane and drive in two modes: first, keep the same lane, and second, overtake back and forth between two lanes with trajectories defined by two parameters of $s$ and $n = {w \mathord{\left/
 {\vphantom {w 2}} \right.
 \kern-\nulldelimiterspace} 2}$ shown in Fig.\ref{fig:synth_scenario}-(a). In the case of changing lanes/overtaking, two values of 2 and 4 seconds are used for $s$ (assuming constant speed). And finally, the lateral offset of the TV start lane from EV's lane varies from -80m to 80m (with a step of 1m), The cyclists' trajectories include only lane-keeping i.e without any lateral motion. There is only one TV in each generated scenario to prevent occlusion, although, in Fig.\ref{fig:synth_scenario} three TVs are depicted which is a combination of three scenarios to be more informative. 

\textcolor{black}{We refer to all synthetic scenarios described above as \textit{primary scenarios}. To further compare the performance metrics against model-free \cite{wang2022detection} and model-based\cite{zhang2017efficient} indirect tracking methods, simulation scenarios designed in \cite{wang2022detection} are replicated (\textit{secondary scenarios}). In these scenarios the TV moves with the speed of 6~m/sec along i) straight right-angled, ii) right turn, and iii) circular paths (see \cite{wang2022detection} for details).}

\subsubsection{KITTI Dataset}
The final evaluation is conducted using the real-world KITTI tracking dataset for multi-object tracking task. Besides the ground truth labels, only LiDAR data from this dataset is used in the current study for the estimation task, however, the colour images of each frame are also used to plot velocity vectors in image coordinate (Fig.\ref{fig:kitti_scenario}-left) using transformation matrix (velo-to-cam). Moreover, since there is no ground truth label for the velocity of objects in the driving environment, it is obtained by tracking the centre of the 3D bounding boxes. In the KITTI dataset, the bounding box coordinates are provided in the camera frame whereas the estimated velocity values are obtained in the LiDAR coordinate system. Therefore, the calculated velocities are transformed to the camera coordinate ($T_{velo-to-cam}$) to calculate estimation error. 

Since we need to compare the estimation results with the ground truth labels, the training sequences are used for velocity and yaw angle error calculations, however, both training and testing sequences are used to obtain detection performance. Adopted from \cite{lee2020moving}, the point cloud data closer than 25m in the lateral direction (left and right), 80m and 15m in longitudinal front and rear directions, respectively are considered, and the rest of the data and labels out of this range are discarded in the evaluation process. 

\begin{figure}[t]
\centering
\includegraphics[width=1\linewidth]{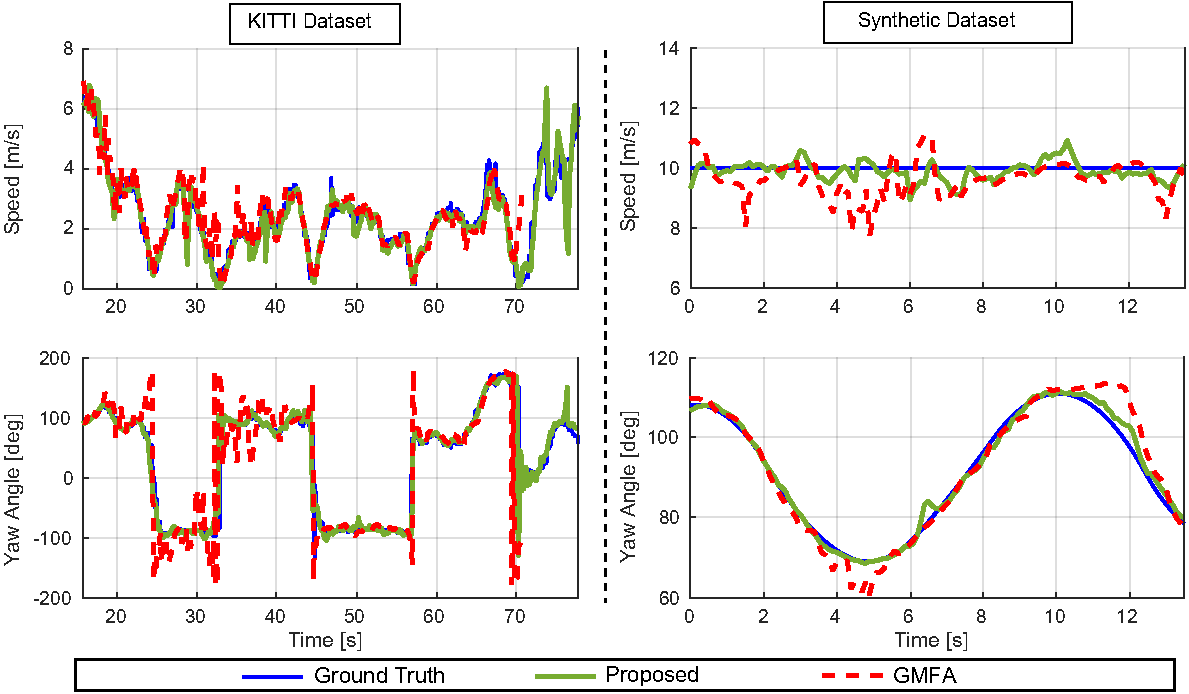}
\caption{Comparative results of the proposed and GMFA algorithms for the synthetic and KITTI datasets}
\label{fig: compare_diagram}
\centering
\end{figure}

\subsection{Evaluation Metrics}
\label{sec: kpis}
Following the previous studies \cite{lee2020moving,lee2022moving}, the velocity vector estimation accuracy is evaluated by speed and angle ($\theta$) errors with respect to the ground truth (GT) values. The estimation errors for $o^{th}$ target object at time $t$ are calculated in Eq.~\ref{eq: kpi_error}.

\begin{equation}
\label{eq: kpi_error}
\begin{array}{l}
\delta v_o^{}\left( t \right) = {\left| {\left\| {{\bf{v}}_o^{GT}} \right\| - \left\| {{\bf{\hat v}}_o^{}} \right\|} \right|_t}\\
\delta \theta _o^{}\left( t \right) = {\left| {\theta _o^{GT} - \hat \theta _o^{}} \right|_t}
\end{array}
\end{equation}

In order to compare DATMO performance throughout all data points, the standard deviation ($\sigma$) of the error distribution of all timesteps and target moving objects is used in Table.\ref{table: result}. Moreover, same as \cite{lee2020moving}, the detection performance is also evaluated by \textit{Precision} and \textit{Recall} defined in Eq.~\ref{eq: kpi_det}. 

\begin{equation}
\label{eq: kpi_det}
    \begin{array}{l}
Pr = \frac{{TP}}{{TP + FP}}\\
Re = \frac{{TP}}{{TP + FN}}
\end{array}
\end{equation}

The last metric to quantify the estimation performance is the time each algorithm takes to process an instance of the LiDAR scan to detect and estimate the state of the moving objects. 

\subsection{Results}
\label{sec: results}
\textcolor{black}{The evaluation results are divided into three sections. Firstly, a stochastic comparative analysis is conducted with model-free direct tracking methods in \cite{lee2020moving} (GMFA) and \cite{li2023high}. Secondly, simulation results compare the proposed method with model-free and model-based indirect tracking methods developed in \cite{wang2022detection} and \cite{zhang2017efficient}. Lastly, the effects of the continuity filters (Section.~\ref{sec: mask}) are presented in an ablation study.}

The grid size in the proposed algorithms is $0.17 \times 0.17\;{\rm{m}}$ and the Farneback optical flow algorithm used in the proposed method is taken from OpenCV computer vision library with the following setting: \textit{NumPyramidLevels = 3}, \textit{PyramidScale = 0.5}, \textit{NumIterations = 3}, \textit{NeighborhoodSize = 3}, and \textit{FilterSize = 11}.

\begin{figure}[b]
\centering
\includegraphics[width=1\linewidth]{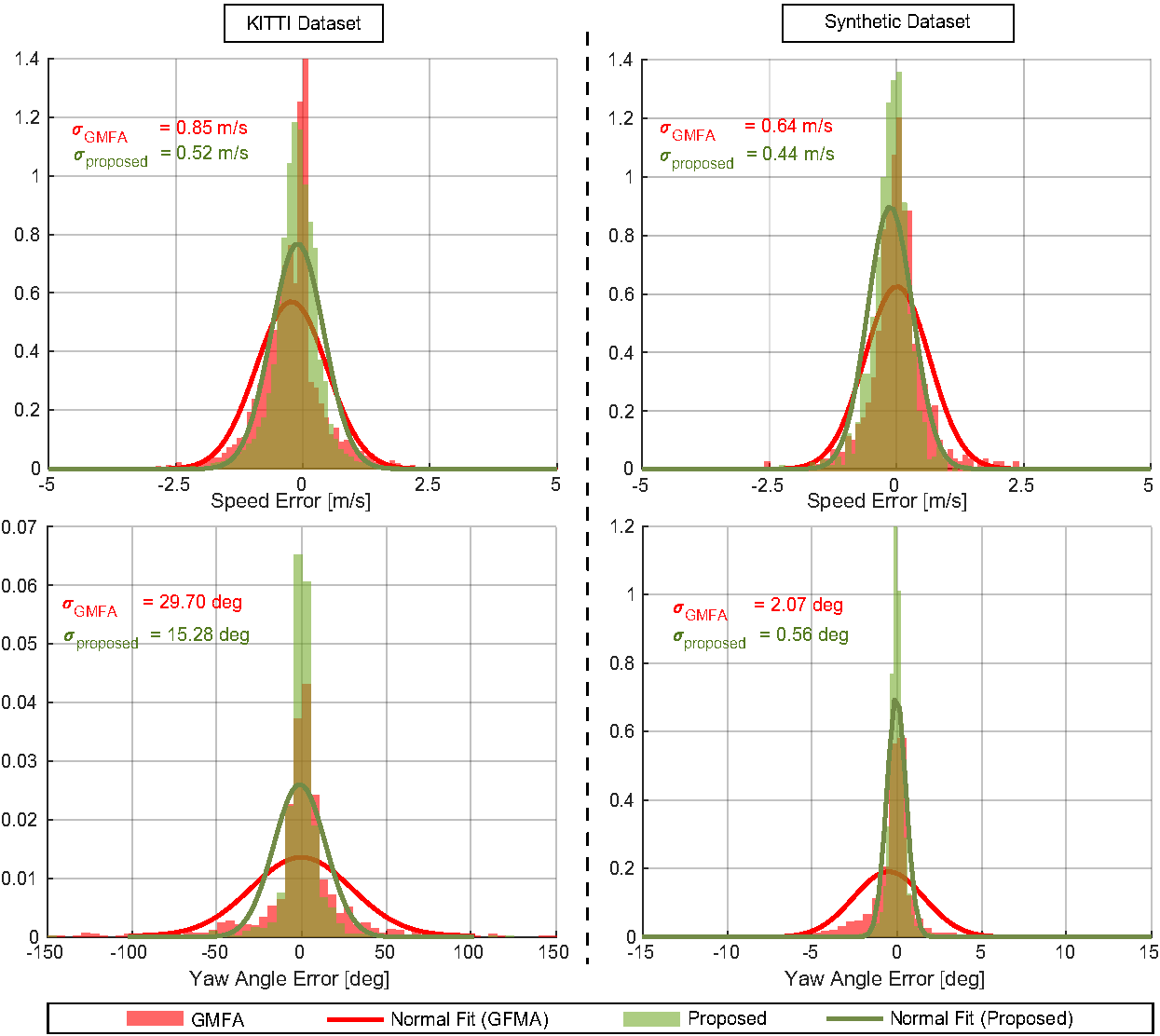}
\caption{Comparative error distribution of the proposed and GMFA \cite{lee2020moving} algorithms for the primary synthetic and KITTI datasets}
\label{fig: compare_dist}
\centering
\end{figure}

\def\ra{.12}
\def\w{1.5}
\begin{table*}
\centering
\caption{Experimental evaluation results of the proposed method and GMFA \cite{lee2020moving} for both synthetic simulation and KITTI dataset}
\begin{tabular}{c>{\centering}m{\w cm}c>{\centering}m{\w cm}>{\centering}m{\w cm}>{\centering}m{\w cm}>{\centering}m{\w cm}m{\w cm}<{\centering}} 
\toprule
\multirow{2}{*}{Dataset}   & $\left\| {\Delta \upsilon } \right\|$   & \multirow{2}{*}{Method} & {Precision} & {Recall}  & ${\sigma _v}$  & ${\sigma _a}$ & ${{\rm{Time}}}$  \\
                            &               ${\left[ {{{\rm{m}} \mathord{\left/
 {\vphantom {{\rm{m}} {\rm{s}}}} \right.
 \kern-\nulldelimiterspace} {\rm{s}}}} \right]}$       &                         &              $(\%)$              &            $(\%)$              &  ${\left[ {{{\rm{m}} \mathord{\left/
 {\vphantom {{\rm{m}} {\rm{s}}}} \right.
 \kern-\nulldelimiterspace} {\rm{s}}}} \right]}$   &    $\left[ {{\rm{deg}}} \right]$  &  $\left[ {{\rm{ms}}} \right]$     \\ 
\midrule
\multirow{4}{*}{Simulation} & \multirow{2}{*}{$ \le 1$} & \Gape[\ra cm] GMFA                    &         \textbf{94.3}                 &       \textbf{93.6}                   &   0.28       &   \textbf{1.12}   &    147   \\ 
\cline{3-8}
                            &                      & \Gape[\ra cm] Proposed                    &          94.1                  &         93.8                &     \textbf{0.23}  &    1.37  &     \textbf{139}  \\ 
\cline{2-8}
                            & \multirow{2}{*}{$ > 1$} & \Gape[\ra cm] GMFA                    &           88.9                 &           \textcolor{black}{87.3}              &      \textcolor{black}{2.28}   &    3.81      &     155  \\ 
\cline{3-8}
                            &                      & \Gape[\ra cm] Proposed                    &          \textbf{89.2}                  &       \textbf{\textcolor{black}{88.3}}                  &    \textbf{\textcolor{black}{1.48}}    &   \textbf{\textcolor{black}{0.63}}   &    \textbf{147}   \\ 
\hline
\multirow{4}{*}{KITTI}      & \multirow{2}{*}{$\le 1$} & \Gape[\ra cm] GMFA                    &      \textbf{89.7}                      &             88.7            &   0.48    &   18.12    &   196    \\ 
\cline{3-8}
                            &                      & \Gape[\ra cm] Proposed                    &          {88.8}                  &          \textbf{89.1}              &  \textbf{0.33 }    &   \textbf{10.97}   &    \textbf{171}   \\ 
\cline{2-8}
                            & \multirow{2}{*}{$ > 1$} & \Gape[\ra cm] GMFA                    &          87.5                  &         86.8                     &   0.78    &   34.21   &   201    \\ 
\cline{3-8}
                            &                      & \Gape[\ra cm] Proposed                    &           \textbf{88.6}                 &              \textbf{88.2}           &    \textbf{0.56}   &  \textbf{16.71}    &    \textbf{183}   \\
\bottomrule
\end{tabular}
\label{table: result}
\end{table*}

\begin{table}
\centering
\caption{Comparing to other model-free and learning-based motion estimation methods}
\begin{tblr}{
  column{-} = {c},
  row{-} = {m},
  column{2} = {1.5cm},
  hlines,
}
Ref & Speed Error [m/sec] & time [ms] & Training needed \\
 Wang et al.\cite{wang2021simultaneous}   & 1.69                & 80        & Yes             \\
 Liu et al.\cite{liu2019flownet3d} & 4.37                & -         & Yes             \\
 Li et al.\cite{li2023high} & 0.42                & 240       & Partially       \\
 Proposed   & 0.44                & 142       & No              
\end{tblr}
\label{table: other_kitti}
\end{table}

\begin{table}
\centering
\caption{Comparing the mean and max estimation errors against model free \cite{wang2022detection} and model-based\cite{zhang2017efficient} indirect tracking methods. Results obtained from the secondary scenarios in \cite{wang2022detection}.}
\begin{tblr}{
  cell{1}{1} = {r=2}{},
  cell{1}{2} = {r=2}{},
  cell{1}{3} = {c=2}{},
  cell{1}{5} = {c=2}{},
  cell{3}{1} = {r=3}{},
  cell{6}{1} = {r=3}{},
  cell{9}{1} = {r=3}{},
  column{-} = {c},
  row{-} = {m},
  column{1} = {.9cm},
  column{3-6} = {.83cm},
  hlines,
}
secondary scenarios& Method       & Speed [m/sec] &      & Direction [deg] &      \\
                   &              & mean          & max  & mean            & max  \\
i                  & Wang et al.\cite{wang2022detection}  & 0.25          & 0.43 & 0.52            & 1.49 \\
                   & Zhang et al.\cite{zhang2017efficient} & 0.38          & 0.59 & 1.23            & 1.93 \\
                   & Proposed     & 0.09         & 0.31 & 0.18            & 0.51 \\
ii                 & Wang et al.\cite{wang2022detection}  & 0.40          & 0.70 & 0.80            & 2.53 \\
                   & Zhang et al.\cite{zhang2017efficient} & 0.52          & 1.00 & 1.53            & 4.02 \\
                   & Proposed     & 0.21          & 0.47 & 0.83            & 1.70 \\
iii                & Wang et al.\cite{wang2022detection}  & 0.29          & 0.40 & 1.65            & 2.50 \\
                   & Zhang et al.\cite{zhang2017efficient} & 0.43          & 0.84 & 2.25            & 5.09 \\
                   & Proposed     & 0.19          & 0.44 & 0.41            & 1.82 
\end{tblr}
\label{table: secondary_simulation}
\end{table}

\textcolor{black}{\subsubsection{Comparison with Direct Tracking Methods}
The} detection and state estimation results for 21 sequences of the KITTI training labelled dataset and more than 800 synthetic driving \textcolor{black}{scenarios (primary)} are presented in this section. In both sets of these datasets, the detection and state estimation performance evaluated for cyclists, sedans, vans (or bigger vehicles such as trucks or buses in the KITTI dataset) and pedestrians are ignored in this study. As a sample, the tracking results of moving object tracking in both KITTI and synthetic datasets have been illustrated in Fig.\ref{fig: compare_diagram} left and right column, respectively, for GMFA and the proposed approaches (dashed red and solid green, respectively). Discontinuation of the dashed red diagram in the left column plot shows that the GMFA couldn't track the object from approximately 71 sec onward. The top row in this figure shows the speed estimations whereas the bottom row depicts the yaw angle estimation results for a moving object in one sequence. The estimated values are reported as relative values i.e measured in the EV coordinate system.

In order to compare the GMFA and the proposed approaches, the estimation error distribution of all sequences for two datasets is obtained. This distribution contains estimation error of all time steps throughout all sequences. Speed and yaw angle estimation error distribution has been shown in Fig.\ref{fig: compare_dist} top and bottom row, respectively. In this figure, the performance of both GMFA (red) and the proposed (green) methods is illustrated for KITTI and synthetic datasets separately in the left and right columns, respectively. Furthermore, for each distribution, a normal distribution function has been fitted with the standard deviation value printed in the top left for both methods using the same colour codes. Similar to \cite{lee2020moving} the standard deviation values are used to compare the accuracy of DATMO methods. 

Finally, the detection and estimation results of two methods and two datasets are summarized in Table.\ref{table: result}. Precision and recall metrics are for evaluating moving object detection while the standard deviation and time columns report the result of the state estimation accuracy and computational cost, respectively. The results are further reported for two different ranges of relative velocity ($\left\| {\Delta \upsilon } \right\|  \le 1$ and $\left\| {\Delta \upsilon } \right\|  > 1{{\;{\rm{m}}} \mathord{\left/
 {\vphantom {{\;{\rm{m}}} {\rm{s}}}} \right.
 \kern-\nulldelimiterspace} {\rm{s}}}$), because the GMFA method of \cite{lee2020moving} is developed for detecting and tracking of moving objects with \textit{``low relative speed''}. Therefore, in order to check if the GMFA method is replicated properly, the estimation errors for low relative speeds ($\left\| {\Delta \upsilon } \right\|  \le 1{{\;{\rm{m}}} \mathord{\left/
 {\vphantom {{\;{\rm{m}}} {\rm{s}}}} \right.
 \kern-\nulldelimiterspace} {\rm{s}}}$) should be less than what was reported in \cite{lee2020moving}. It should be noted that since there is no exact velocity ground truth label for the KITTI tracking dataset, the calculated error for this dataset even for low speeds is not comparable directly with values reported in \cite{lee2020moving}, and we use the replicated GMFA algorithm instead to only compare the final estimation error with the proposed approach performance. Moreover, the processing time reported in this table is the average time the computing unit (Intel Core(TM) i7-7600 CPU @ 2.80GHz) needs for each cycle excluding the first step of each sequence which needs extra initialization time. \textcolor{black}{The breakdown of computational complexity for different processes within the framework is given in Table.~\ref{table: time}. Since we believe that the core process of optical flow in the proposed method could be parallelized using off-the-shelf tools, this process was implemented on both CPU and GPU (GeForce RTX 2080 Ti) for the simulation scenarios. The results indicate an 80\% improvement in processing time for GPU compared to CPU.}

\textcolor{black}{The estimation error and computational complexity comparison with other model-free and learning-based motion estimation methods is summarised in Table.~\ref{table: other_kitti}. The performance metrics of other methods and evaluation conditions are adopted from \cite{li2023high}. The results are based on the KITTI tracking dataset, sequences 0000, 0005, and 0010. The objects within a radius of 50 m are considered. The results, represented in Table.~\ref{table: other_kitti}, include other learning-based motion estimation methods as additional references. The comparison suggests that although our method's performance in terms of speed accuracy is comparable to that of \cite{li2023high}, there is a significant improvement in computational complexity. This is attributed to the iterative nature of the ICP method used in \cite{li2023high}. Moreover, all other methods in this table are data-driven and will need retraining the different situation or sensor configurations; otherwise, the performance may decline. While our method is deterministic and does not need training.
}

\textcolor{black}{\subsubsection{Comparison with Indirect Tracking Methods}
Table \ref{table: secondary_simulation} presents secondary simulation results for comparison with indirect tracking methods. Like the simulation scenarios used in this comparison, the performance metrics and values for other methods in this table are sourced from \cite{wang2022detection}. As anticipated, the proposed method outperforms both indirect tracking methods, as they compute velocity based on macroscopic classified point clouds. In contrast, our framework employs the optical flow algorithm to calculate the velocity field at a microscopic level (grid-based) before the classification and EKF tracking processes take place.
}

\begin{figure}[!b]
\centering
\includegraphics[width=1\linewidth]{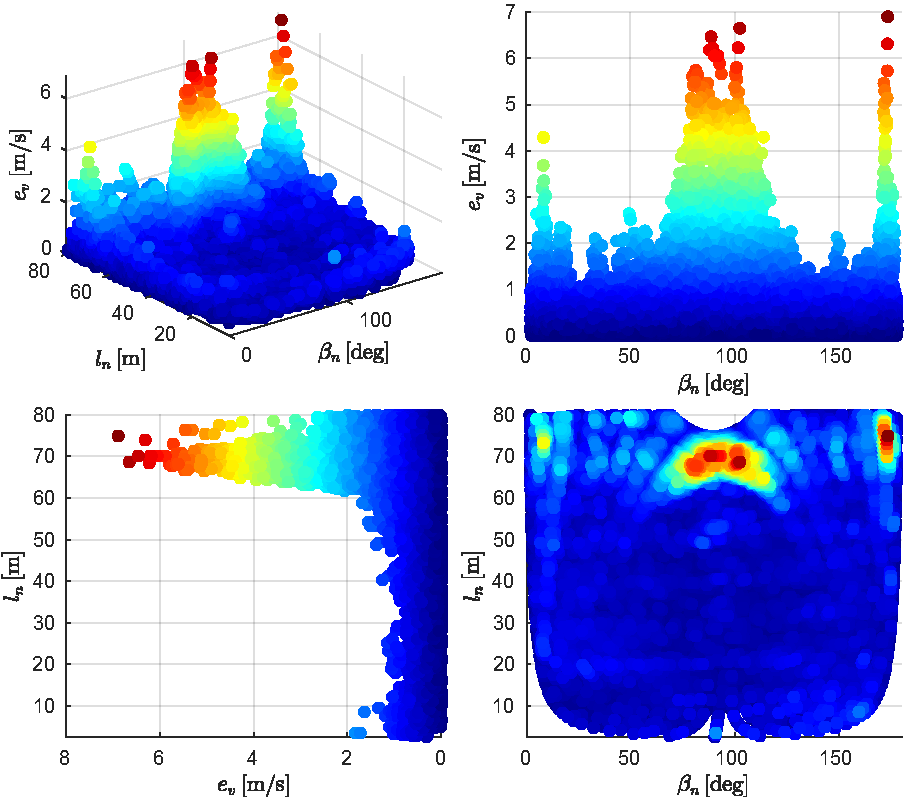}
\caption{Error model as a function of orientation and distance of the TV with respect to EV for $\Delta \upsilon  = 10\;{{\rm{m}} \mathord{\left/
 {\vphantom {{\rm{m}} {\rm{s}}}} \right.
 \kern-\nulldelimiterspace} {\rm{s}}}$. The heatmap colour corresponds to the absolute error value in ${e_\upsilon }$ axis}
\label{fig: errModel}
\centering
\end{figure}

\textcolor{black}{
\subsubsection{Ablation Study}
The impact of the two-layer filters applied to the velocity vector field and ablation experiment is investigated in this section. The \textit{continuity} and \textit{propagation} filters (see Section.~\ref{sec: mask}) are disabled individually to quantify their effect through the changes in the performance metrics. Only the KITTI dataset is used for this experiment and the results are summarized in Table.~\ref{table: ablation}
\\
Based on the findings presented in Table.~\ref{table: ablation}, the performance metrics show improvement with the inclusion of both filters, with notably significant enhancement attributed to the propagation filter.}

\textcolor{black}{\subsection{Estimation Error Sensitivity to TV's Configuration}
After} validating the proposed DATMO approach and comparing it with the state-of-the-art method, the sensitivity of the speed estimation error to the changes in the TV's configuration for the proposed method is explored in this section. This would help other researchers who use this tracking method (motion planning and control) to consider an error model. Two elements of the TV configuration (${\beta _n},{l_n}$) are used as variables in this section. In other words, we want to investigate how the estimation error changes by changing the distance (${l_n}$) and orientation (${\beta _n}$) of the TV. The synthetic data is used to sweep these variables and the proposed algorithm is applied to measure the estimation error for each case. The result of this sensitivity analysis is presented in a 3D plot and three 2D plots (three views of the same 3D plot) in Fig.\ref{fig: errModel}. In this figure, the heatmap colour correlated to the absolute speed estimation error (${e_\upsilon }$) is used to visualize the error value, particularly in the top-view plot (red and blue colours corresponding to high and low absolute error, respectively).

\section{Discussion}
\label{sec: discussion}
The obtained results are presented in the section.\ref{sec: results} are further discussed in detail in this section. The comparative data reported in Table\ref{table: result} and Fig.\ref{fig: compare_dist} show the superior performance of the proposed method compared with the GMFA approach. But before comparing the two approaches, we need to validate the regenerated algorithm for GMFA. Since this algorithm is originally developed for low relative speed and has been validated with an autonomous vehicle platform, the performance of the regenerated algorithm for low-speed synthetic dataset should surpass the values reported in \cite{lee2020moving} (the standard deviation of the speed and yaw angle error are 0.40 m/s and 1.81 deg, respectively). According to Table.\ref{table: result} (first row), the GMFA result for low-speed simulation outperforms these outcomes. Therefore, the regenerated GMFA algorithm is reliable to be tested as the baseline with other datasets such as KITTI or high relative speed synthetic datasets.

In detecting the moving objects, precision and recall values show almost similar performance for both comparing methods (increased only $1\%$ in the proposed approach). However, the state estimation accuracy shows more than $34\%$ and $50\%$ improvement in the standard deviation of the estimated speed and yaw angle error distribution, respectively. The fitted normal distribution along with the standard deviations for both synthetic and KITTI datasets has been shown in Fig.\ref{fig: compare_dist}. 

\begin{table}[t]
\centering
\caption{Ablation study for the two-layer filter applied on velocity vector field}
\begin{tblr}{
column{1-5} = {c},
row{1-4} = {m},
hlines,
column{2,3}={c}{1.2cm},
}
 \# & Propagation Filter & Continuity Filter & ${\sigma _a}$ [m/sec] & ${\sigma _a}$ [deg] \\
1 &      \xmark        &       \xmark        &    2.11   &   0.93  \\
2 &      \cmark        &       \xmark        &     1.45  &   0.70  \\
3 &      \cmark        &       \cmark        &    1.38   &  0.64   
\end{tblr}
\label{table: ablation}
\end{table}

\begin{table}
\centering
\caption{Processing time for different components of the proposed method}
\begin{tblr}{
  column{-} = {c},
  row{-} = {m},
  column{3,4,6} = {1.2cm},
  column{2,5} = {.7},
  hlines, 
}
\textbf{Process:}   & Data Parsing & 3D to 2D Conversion & Optical Flow (GPU/CPU) & GNN Tracker & Total (GPU/CPU) \\
\textbf{Time [ms]:} & 8.1           & 4.4                  & 4.6/119           & 10.8         & 27.9/142.3    
\end{tblr}
\label{table: time}
\end{table}

Moreover, the measured processing times show an average of $\sim 8 \%$ improvement for the proposed method compared with the GMFA. The computation time for the synthetic data shows less improvement compared to the KITTI dataset ($5\%$ vs $10.5\%$). Since each sequence of the synthetic scenario contains only one moving object, and the GMFA is based on point cloud registration for which the processing time depends on the number of detected moving point clusters, the computational effort is more consistent and less than that of KITTI scenarios in which the number of moving objects are more than one in most sequences. All computations in this study were done on CPU without parallel processing, whereas the proposed method which is based on an optical flow algorithm, has the potential to be implemented on GPU to accelerate the computations. This is another advantage of this approach over point cloud registration-based methods such as GMFA that use optimization and this makes it more difficult to parallelize the computations. Recently, Contemporary GPUs dedicate hardware to particularly accelerate optical flow algorithms up to 10 times faster \cite{Aruna2019}. Therefore, using parallel computation will accelerate the processing even more for the proposed approach.  

Overall, the comparison results indicate that the proposed method's performance in state estimation and computational cost is comparable with the state-of-the-art method (GMFA), and as the last part of analysing the results error sensitivity of the proposed method is considered. The estimation error sensitivity to the configuration of the TV, illustrated in Fig.\ref{fig: errModel}, shows that the error magnitude is more sensitive to the orientation of the TV when the target vehicle is located at farther distances (${l_n} > 45\;{\rm{m}}$). The way that the error value changes with respect to the orientation of the TV ($\beta_n$) is also interesting. The error increases at three specific orientations: ${\beta _n} = 0,90,180\;\deg $. regarding Fig.\ref{fig:dynamics}, the first (${\beta _n} = 0\;\deg $) and last (${\beta _n} = 180\;\deg $) orientations correspond to the configuration in which TV facing or backing on to the EV, whereas the second orientation (${\beta _n} = 90\;\deg $) is for the case in which TV's side is toward the EV i.e LiDAR sensor location. One of the possible reasons for this correlation could be the fact that in these configurations the scanned point cloud is no longer scattered in 3D space and mostly on a 2D plane. For instance, in ${\beta _n} = 90\;\deg $ most of the scanned points are from the side of the vehicle. However, to elaborate more on the error model and consider all involved factors more research is required in future studies.

\section{Conclusion}
\label{sec: conclusion}
In this study, a novel DATMO technique was proposed using a Farneback optical flow algorithm. This study revealed the promising potential of this approach in terms of accuracy and processing costs. Similar to traditional GMFA techniques, the optical-flow-based technique approach proposed and studied in this paper demonstrated good resilience against variations of the object sizes in driving scenes. Analysis of the error sensitivity to the configuration of the target vehicle in this study revealed meaningful correlations which could be used in future for error modelling. Our results showed that the error values increase when the TV moving in radial (${\beta _n} = 0,180\;\deg $) and tangential (${\beta _n} = 90\;\deg $) directions in distances farther than $50\;{\rm{m}}$. It shall be noted that Small size objects such as pedestrians were not covered in our study. Further studies could explore estimating the state of pedestrians by reducing the grid size and implementing the algorithm using parallel computing to calculate optical flow.

\section*{Acknowledgment}
This research is sponsored by the Centre for Doctoral Training to Advance the Deployment of Future Mobility Technologies (CDT FMT) at the University of Warwick.

\appendix
\section*{Deriving Angular velocity from velocity vector field}
\label{app: curl}
\begin{figure}[!t]
\centering
\includegraphics[width=0.6\linewidth]{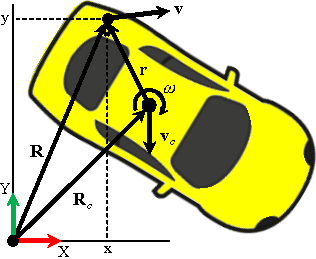}
\caption{Velocity of points on moving rigid body (vehicle)}
\label{fig:curlAppendix}
\centering
\end{figure}

\noindent Assuming rigid body motion, the angular velocity could be obtained from the velocity vector field. If $\left\{ {\hat i,\hat j,\hat k} \right\}$ are unit vectors in $\left\{ {x,y,z} \right\}$, respectively, and considering notation used in Fig.~\ref{fig:curlAppendix}, the angular velocity for planar motion is derived as follows:
\begin{equation*}
\begin{array}{*{20}{l}}
{\bf{v}}& = &{{{\bf{v}}_c} + {\bf{\omega }} \times {\bf{r}}}\\
{}& = &{{{\bf{v}}_c} + {\bf{\omega }} \times \left( {{\bf{R}} - {{\bf{R}}_c}} \right)}\\
{}& = &{\left( {{{\bf{v}}_c} - {\bf{\omega }} \times {{\bf{R}}_c}} \right) + {\bf{\omega }} \times {\bf{R}}}
\end{array}
\end{equation*}

\noindent Rewriting this equation by substituting ${\bf{R}} = x{\bf{\hat i}} + y{\bf{\hat j}}$, and ${{\bf{V}}_c} = {{\bf{v}}_c} - {\bf{\omega }} \times {{\bf{R}}_c} = {V_{cx}}{\bf{\hat i}} + {V_{cy}}{\bf{\hat j}}$:

\begin{equation*}
\begin{array}{*{20}{l}}
{\bf{v}}& = &{{{\bf{V}}_c} + \left( { - \omega y{\bf{\hat i}} + \omega x{\bf{\hat j}}} \right)}\\
{}& = &{\left( {{V_{cx}} - \omega y} \right){\bf{\hat i}} + \left( {{V_{cy}} + \omega x} \right){\bf{\hat j}}}
\end{array}
\end{equation*}

\noindent And by applying the curl operator to both sides, the angular velocity is obtained based on the curl of the vector field ${\bf{v}}$. It should be noted that the rigid body assumption makes the curl independent of the position and linear velocity of the centre $c$.
\begin{equation*}
\begin{array}{l}
\begin{array}{*{20}{l}}
{\nabla  \times {\bf{v}}}& = &{\left[ {\frac{\partial }{{\partial x}}\left( {{V_{cy}} + \omega x} \right) - \frac{\partial }{{\partial y}}\left( {{V_{cx}} - \omega y} \right)} \right]{\bf{\hat k}}}\\
{}& = &{2\omega {\bf{\hat k}}}
\end{array}\\
\begin{array}{*{20}{l}}
 \Rightarrow &\omega & = & 0.5\left( {\nabla  \times {\bf{v}}} \right)
\end{array}
\end{array}
\end{equation*}

\bibliographystyle{myIEEEtran}
\bibliography{IEEEabrv,bibliography}

\vskip 0pt plus -1fil

\begin{IEEEbiography}[{\includegraphics[width=1in,height=1.25in,clip,keepaspectratio]{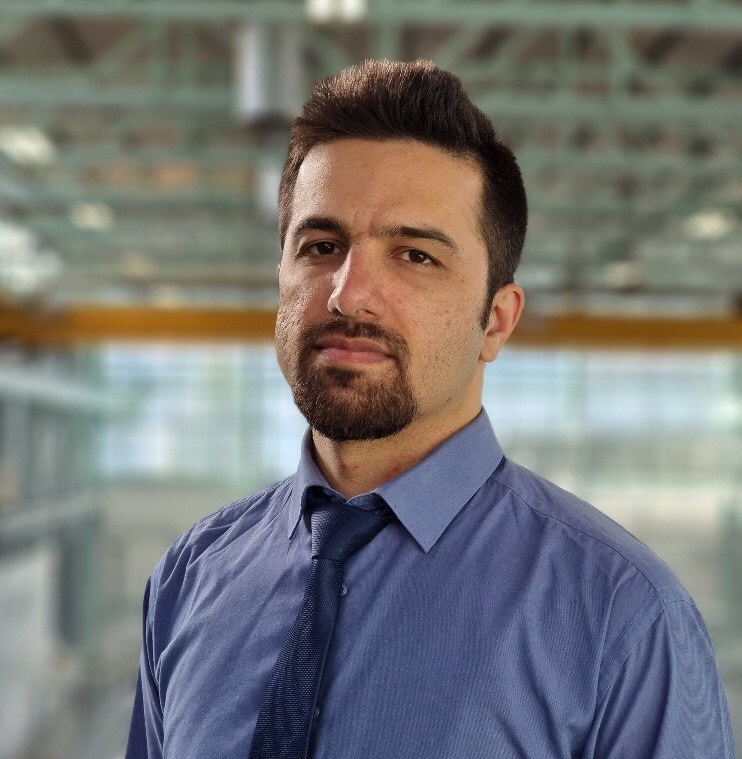}}]{Mohammadreza Alipour Sormoli}
received the M.Sc. degree from the Amirkabir University of Technology (Tehran Polytechnic) in 2017. worked as a research assistant at Koc University and is currently working toward the PhD degree in autonomous driving technology at the University of Warwick (WMG). His research interests include robotics, mechatronics, control and dynamics of autonomous systems.
\end{IEEEbiography}

\begin{IEEEbiography}[{\includegraphics[width=1in,height=1.25in,clip,keepaspectratio]{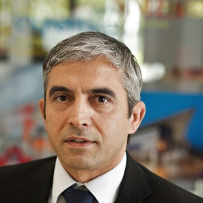}}]{Mehrdad Dianati}
(Senior Member, IEEE) is a professor of connected and cooperative autonomous vehicles at WMG, the University of Warwick and the School of EEECS at the Queen's University of Belfast. He has been involved in a number of national and international projects as the project leader and the work-package leader in recent years. Prior to academia, he worked in the industry for more than nine years as a Senior Software/Hardware Developer and the Director of Research and Development. He frequently provides voluntary services to the research community in various editorial roles; for example, he has served as an Associate Editor for the IEEE Transactions On Vehicular Technology.  He is the Field Chief Editor of Frontiers in Future Transportation.
\end{IEEEbiography}

\begin{IEEEbiography}[{\includegraphics[width=1in,height=1.25in,clip,keepaspectratio]{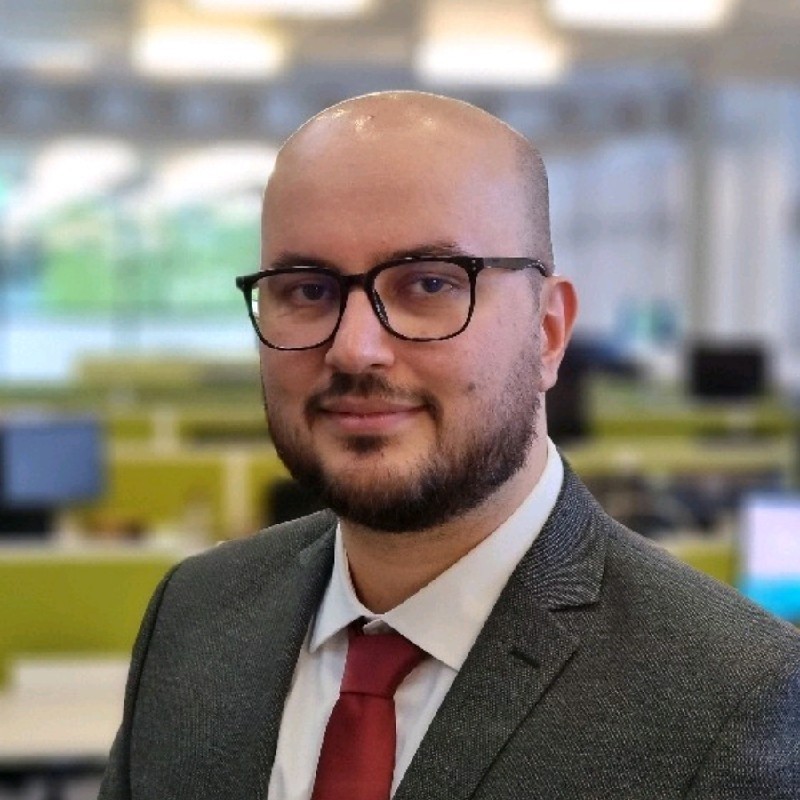}}]{Sajjad Mozaffari} received the B.Sc. and M.Sc. degrees in electrical engineering from the University of Tehran, Tehran, Iran, in 2015 and 2018, respectively. He is currently working toward the PhD degree with the Warwick Manufacturing Group, University of Warwick, Coventry, U.K. His research interests include machine learning, computer vision, and connected and autonomous vehicles.
\end{IEEEbiography}

\begin{IEEEbiography}[{\includegraphics[width=1in,height=1.25in,clip,keepaspectratio]{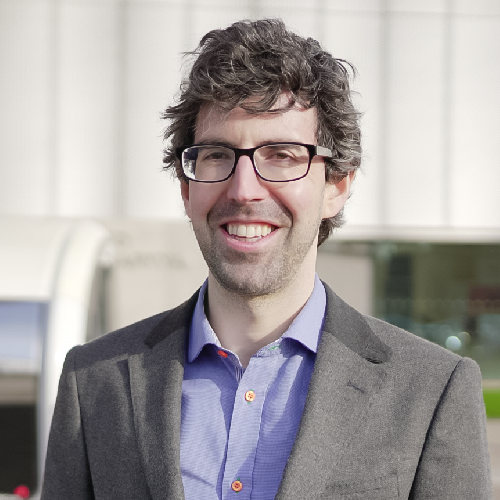}}]{Roger Woodman}
is an Assistant Professor and Human Factors research lead at WMG, University of Warwick. He received his PhD from Bristol Robotics Laboratory and has more than 20 years of experience working in industry and academia. Among his research interests, are trust and acceptance of new technology with a focus on self-driving vehicles, shared mobility, and human-machine interfaces. He has several scientific papers published in the field of connected and autonomous vehicles. He lectures on the topic of Human Factors of Future Mobility and is the Co-director of the Centre for Doctoral Training, training doctoral researchers in the areas of intelligent and electrified mobility systems.
\end{IEEEbiography}

\end{document}